%% file: emnlp2020.tex
\documentclass[11pt,a4paper]{article}
\usepackage[hyperref]{emnlp2020}
\hypersetup{
    pdfsubject={A Fully Hyperbolic Neural Model for Hierarchical Multi-Class Classification},
    pdftitle={A Fully Hyperbolic Neural Model for Hierarchical Multi-Class Classification},
    pdfauthor={Federico Lopez},
    pdfkeywords={natural language processing, hyperbolic space. hierarchical multiclass classification, fine-grained entity typing, entity typing, hierarchy}
}
\usepackage{times}
\usepackage{latexsym}

\usepackage{microtype}

\usepackage{amsmath} % for \mathcal
\usepackage{amssymb}
\usepackage[disable]{todonotes}  % \todo{note}
\usepackage{booktabs} % for \specialrule command
\usepackage{graphicx}
\usepackage{subfig}
\graphicspath{{./fig/}{./fig_raw/}}
\usepackage{booktabs}
\usepackage{multirow}
\usepackage{adjustbox}
\usepackage{mathtools} % for multiline inside equation
\usepackage{textcomp}  % for \textsection

\usepackage{tabulary} % to set table width

\DeclareUnicodeCharacter{2212}{-} % to use minus in the appendix

\usepackage{url}

\aclfinalcopy % Uncomment this line for the final submission
\def\aclpaperid{490} %  Enter the acl Paper ID here

\newcommand{\fone}{$\operatorname{F}_1$}
\newcommand{\manifold}{$\mathcal{M}$}
\newcommand{\real}{$\mathbb{R}$}
\newcommand{\realto}[1]{$\mathbb{R}^{#1}$}

\newcommand{\modbase}{\textsc{base}}
\newcommand{\modlarge}{\textsc{large}}
\newcommand{\modxlarge}{\textsc{xLarge}}

\title{A {F}ully {H}yperbolic {N}eural {M}odel \\ for {H}ierarchical {M}ulti-{C}lass {C}lassification}

\author{Federico L\'opez \hspace{3cm} Michael Strube \\
    Heidelberg Institute for Theoretical Studies \\
    Research Training Group AIPHES \\
  {\tt firstname.lastname@h-its.org}
}

\begin{document}
\maketitle
\begin{abstract}
Label inventories for fine-grained entity typing have grown in size and complexity. Nonetheless, they exhibit a hierarchical structure. Hyperbolic spaces offer a mathematically appealing approach for learning hierarchical representations of symbolic data. However, it is not clear how to integrate hyperbolic components into downstream tasks.
This is the first work that proposes a fully hyperbolic model for multi-class multi-label classification, which performs all operations in hyperbolic space. We evaluate the proposed model on two challenging datasets and compare to different baselines that operate under Euclidean assumptions. 
Our hyperbolic model infers the latent hierarchy from the class distribution, captures implicit hyponymic relations in the inventory, and shows performance on par with state-of-the-art methods on fine-grained classification with remarkable reduction of the parameter size.
A thorough analysis sheds light on the impact of each component in the final prediction and showcases its ease of integration with Euclidean layers. \footnote{Code available at:\\ \url{https://github.com/nlpAThits/hyfi}}
\end{abstract}

%%%%%%%%%%%%%%%%%%%%%%%%%%%%%%%%%%%%%%%%%%%%%%%%%%%%%%%%%%%%%%%%%%%%%%
\section{Introduction}
% 1 - What problem are you solving?
Entity typing classifies textual mentions of entities, according to their semantic class, within a set of labels (or classes) organized in an inventory.
%Multi-label text classification is the task of assigning to a sample all the relevant labels from a label (or class) inventory \cite{sillaFreitas2011hierclassif}.
The task has progressed from recognizing a few \textit{coarse} classes \cite{sang2003conll}, to extremely large inventories, with hundreds \cite{gillick2014context} or thousands of labels \cite{choi2018ultra}.
Therefore, exploiting inter-label correlations has become critical to improve performance.

\begin{figure}[!t]
%\vspace{-3mm}
\centering
\includegraphics[width=\linewidth,keepaspectratio]{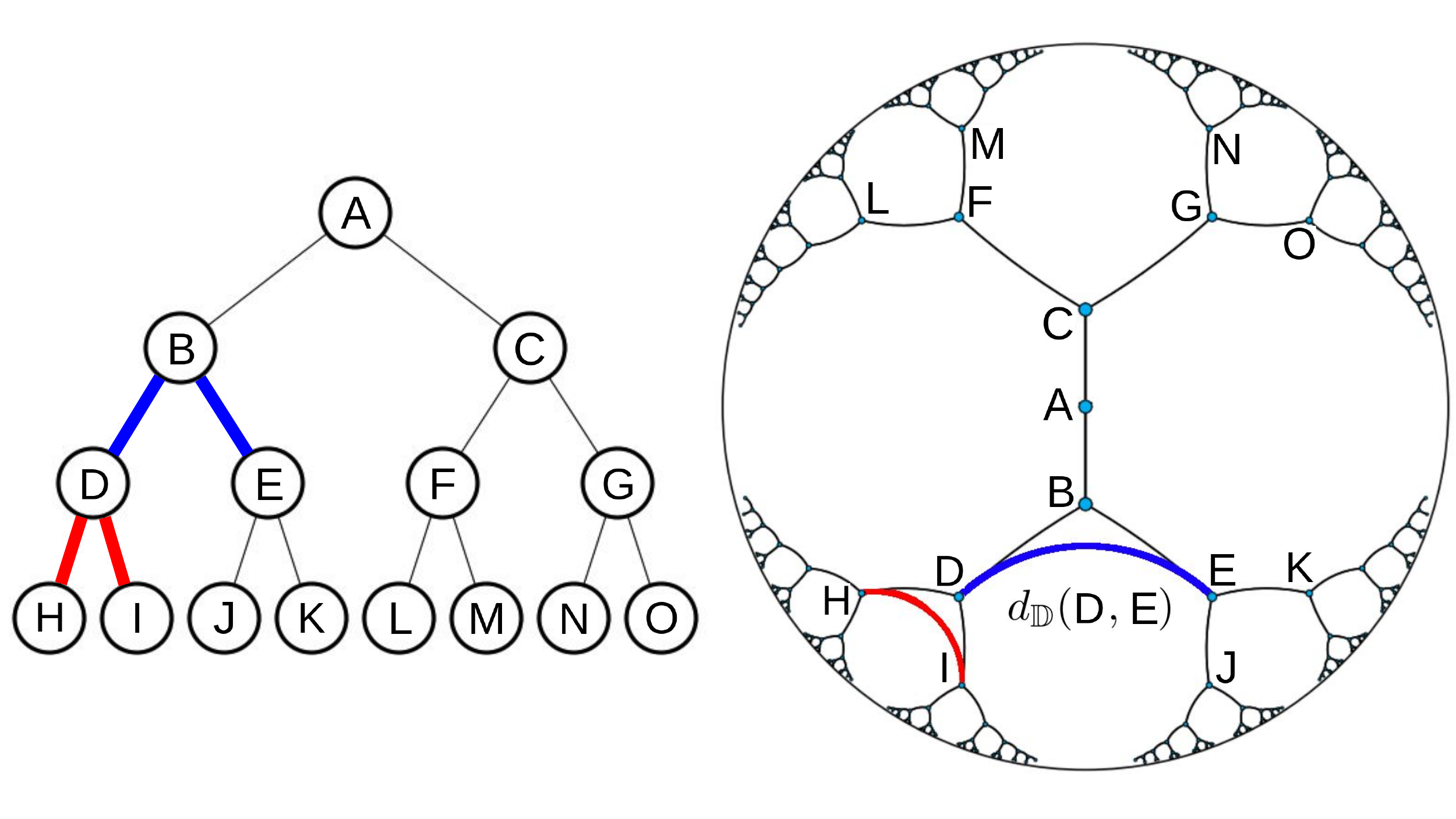}
%\vspace{-7mm}
\caption{Tree embedded in hyperbolic space. %All connected nodes keep the same hyperbolic distance from each other (see Eq.\ \ref{eq:hyper-dist}). 
Items at the top of the hierarchy are placed near the origin of the space, and lower items near the boundary.
Moreover, %$\operatorname{D}$ and $\operatorname{E}$ 
the hyperbolic distance (Eq.\ \ref{eq:hyper-dist}) between sibling nodes resembles the one through the common ancestor, analogous to the distance in the tree. %$\operatorname{B}$.
That is $d_{\mathbb{D}}(\operatorname{D}, \operatorname{E}) \approx d_{\mathbb{D}}(\operatorname{D}, \operatorname{B}) + d_{\mathbb{D}}(\operatorname{B}, \operatorname{E})$.}
\label{fig:cover}
%\vspace{-4mm}
\end{figure}

% 2 - Why is it an interesting/important problem?
% es interesante porque son buenos para modelar redes y estructuras jerárquicas.
% Problema: su adopcion en nlp ha sido baja dado que no hay una forma muy intuitiva de modelar texto en ellos. Distintos papers muestran como agregar un pequeño cambio pero no una aplicacion real y completa
Large inventories tend to exhibit a hierarchical structure, either by an explicit tree-like arrangement of the labels (\textit{coarse} labels at the top, \textit{fine-grained} at the bottom), or implicitly through the label distribution in the dataset (\textit{coarse} labels appear more frequently than \textit{fine-grained} ones).
%A natural solution for dealing with large inventories is to organize them in hierarchy ranging from general, \textit{coarse} labels near the top, to more specific, \textit{fine} classes at the bottom.
Prior work has integrated only explicit hierarchical information by formulating a hierarchy-aware loss \cite{ murty2018hierarchicaLosses, xuBarbosa2018hierarchyAware} or by representing instances and labels in a joint Euclidean embedding space \cite{shimaoka2017neural, abhishek2017jointLearning}. 
However, the resulting space is hard to interpret, and these methods fail to capture implicit relations in the label inventory.
Hyperbolic space is naturally equipped for embedding symbolic data with hierarchical structures \cite{nickel2017poincare}.
Intuitively, that is because the amount of space grows exponentially as points move away from the origin. This mirrors the exponential growth of the number of nodes in trees with increasing distance from the root \cite{cho2019largeMarginClassif} (see Figure~\ref{fig:cover}).
%Its tree-like properties make it efficient to learn hierarchical representations with low distortion \cite{deSa18tradeoffs}. 

% Embeddings  that  are  close  to  the  origin  of  the  disk  will have a relatively small distance to all other points, rep-resenting the root of the hierarchy.  On the other hand,embeddings that are close to the boundary of the disk will have a relatively large distance to all other points and are well suited to represent leaf nodes

% 3 - How are you going to solve it?
In this work, we propose a fully hyperbolic neural model for fine-grained entity typing. Noticing a perfect match between hierarchical label inventories in the linguistic task and the benefits of hyperbolic spaces, we endow a classification model with a suitable geometry to capture this fundamental property of the data distribution. By virtue of the hyperbolic representations, the proposed approach automatically infers the latent hierarchy arising from the class distribution and achieves a meaningful and interpretable organization of the label space. This arrangement captures implicit hyponymic relations (\textit{is-a}) in the inventory and enables the model to excel at fine-grained classification. To the best of our knowledge, this work is the first to apply hyperbolic geometry from beginning to end to perform multi-label classification on real NLP datasets.

%NICE PHRASE FROM GULCEHRE: The focus of this work is to endow neural network representations with suitable geometry to capture fundamental properties of data... given the perfect fit between the label distribution in the linguistic task of entity typing and the mathematical properties of hyperbolic spaces.

% esto deberia ser "hay componentes ya hechos". Y lo conecto al toque con el parrafo sig.

Recent work has proposed hyperbolic neural components, such as word embeddings \cite{tifrea2018poincareGlove}, recurrent neural networks \cite{ganea2018hyperNN} and attention layers \cite{gulcehre2018hyperAttentionNet}.
%Advantages of hyperbolic representations are well-established for discrete data such as networks \cite{krioukov2010hypernetworks} and graphs \cite{liu2019hypergraphsnn, chami2019hgcnn}. In the realm of Natural Language Processing (NLP) components that exploit hyperbolic geometry have been developed as well, such as word embeddings \cite{dhingra2018embeddingTextInHS, tifrea2018poincareGlove}, recurrent neural networks \cite{ganea2018hyperNN} and attention layers \cite{gulcehre2018hyperAttentionNet}.
%or classifiers \cite{cho2019largeMarginClassif} Me encanta este paper pero no hace NLP :(. 
However, researchers have incorporated these isolated components into neural models, whereas the rest of the layers and algorithms operate under Euclidean assumptions.
%\footnote{For a discussion of the term "Euclidean" in the context of Deep Learning see Appendix~\ref{sec:appendix-euclideannn}}.
This impedes models from fully exploiting the properties of hyperbolic geometry. Furthermore, there are different analytic models of hyperbolic space, and not all previous work operates in the same one, which hinders their combination, and hampers their adoption for downstream tasks (e.g. \newcite{tifrea2018poincareGlove} learn embeddings in the Poincar\'e model, \newcite{gulcehre2018hyperAttentionNet} aggregate points in the Klein model, or \newcite{nickel2018lorentz} perform optimization in the Lorentz model).
We address these issues. Our model encodes textual inputs, applies a novel attention mechanism, and performs multi-class multi-label classification, executing all operations in the Poincar\'e model of hyperbolic space (\textsection \ref{sec:model}).
We evaluate the model on two datasets, namely Ultra-Fine \cite{choi2018ultra} and OntoNotes \cite{gillick2014context}, and compare to Euclidean baselines as well as to state-of-the-art methods for the task \cite{xiong2019inductiveBias, onoe-durrett-2019-denoise}. The hyperbolic system % performs on par with the model of \newcite{xiong2019inductiveBias} and 
has competitive performance when compared to an ELMo model \cite{peters2018elmo} and a BERT model \cite{devlin-etal-2019-bert} on very fine-grained types, with remarkable reduction % of $85\%$ and $96\%$ 
of the parameter size %, respectively 
(\textsection\ref{sec:results-and-discussion}).
%\todo[inline]{Cambiar esta frase a la idea de que "imponer the right metric es como imponer the right bias"} 
%We impose an inductive bias on the model by means of the geometry of its internal representation. This allows us to operate on very low-dimensional spaces thus substantially reducing the parameter cost.
Instead of relying on large pre-trained models, we impose a suitable inductive bias by choosing an adequate metric space to embed the data, which does not introduce extra burden on the parameter footprint.
%Phrase from xiong2019inductiveBias: "Instead of using an explicit graphical model, we enforce a relational bias on model parameters, which does not introduce extra burden on label decoding."
% Misma idea pero yo meto el bias en la representacion, lo cual no introduce un costo adicional y permite operar con MUCHOS menos parámetros.

%Our components are developed in a modular way which allows them to be seamlessly integrated into NLP architectures. 

%\todo{Remove!}{While there now exist several hyperbolic components, a practitioner faced with these options has a simple question: How to integrate them with conventional layers? In this work, we answer this question.} 
By means of the exponential and logarithmic maps (explained in \textsection\ref{sec:background}) we are able to mix hyperbolic and Euclidean components into one model, aiming to exploit their strengths at different levels of the representation.
We perform a thorough ablation that allows us to understand the impact of each hyperbolic component in the final performance of the system (\textsection\ref{sec:embed-ablation} and \textsection\ref{sec:system-ablation}), and showcases its ease of integration with Euclidean layers.

%In summary, we make the following contributions:
%%%%%\vspace{-3mm}
%\begin{itemize}
%\itemsep0em 
%    \item We propose a fully hyperbolic neural model for multi-class multi-label classification that performs all operations in hyperbolic space.
%    \item We evaluate the hyperbolic model on two datasets for fine-grained entity typing, with consistent results over Euclidean baselines, and competitive performance with state-of-the-art systems.
%    \item We perform a thorough ablation of the hyperbolic model, which sheds light on the impact of each component in the final prediction, and showcases its ease of integration with Euclidean layers.
%\end{itemize}

%%%%%%%%%%%%%%%%%%%%%%%%%%%%%%%%%%%%%%%%%%%%%%%%%%%%%%%%%%%%%%%%%%%%%%
\section{Hyperbolic Neural Networks}
\label{sec:background}

% intro de la seccion que dice que aca explicamos las operaciones basicas que van a ser usadas en el resto del paper.
%At the heart of neural networks, there are fundamental mathematical operations, such as
%\todo[inline]{Aca (background) y en el cuerpo del modelo es donde mas tengo que achicar el paper. No me puede consumir tanto partes que no son lo que yo hice} While vector addition, matrix-vector multiplication, and pointwise non-linearities are well-known in Euclidean space, their counterparts in hyperbolic space are non-trivial. 
In this section we briefly recall the necessary background on hyperbolic neural components. The terminology and formulas used throughout this work follow the formalism of M\"obius gyrovector spaces \cite{ungar2008hypergeomAndEinstein, ungar2008gyrovector}, and the definitions of hyperbolic neural components of \newcite{ganea2018hyperNN}. For more information about Riemannian geometry and M\"obius operations see Appendix \ref{sec:appendix-riemmanian} and \ref{sec:appendix-mob-operations}. In the following, $\langle \cdot , \cdot \rangle$ and $\| \cdot \|$ are the inner product and norm inherited from the Euclidean space.

\begin{figure}[!t]
%\vspace{-3mm}
\centering
\subfloat{\label{fig:mob-add}{\includegraphics[clip, trim=2.5cm 0.8cm 3cm 1.2cm, width=.5\linewidth,keepaspectratio]{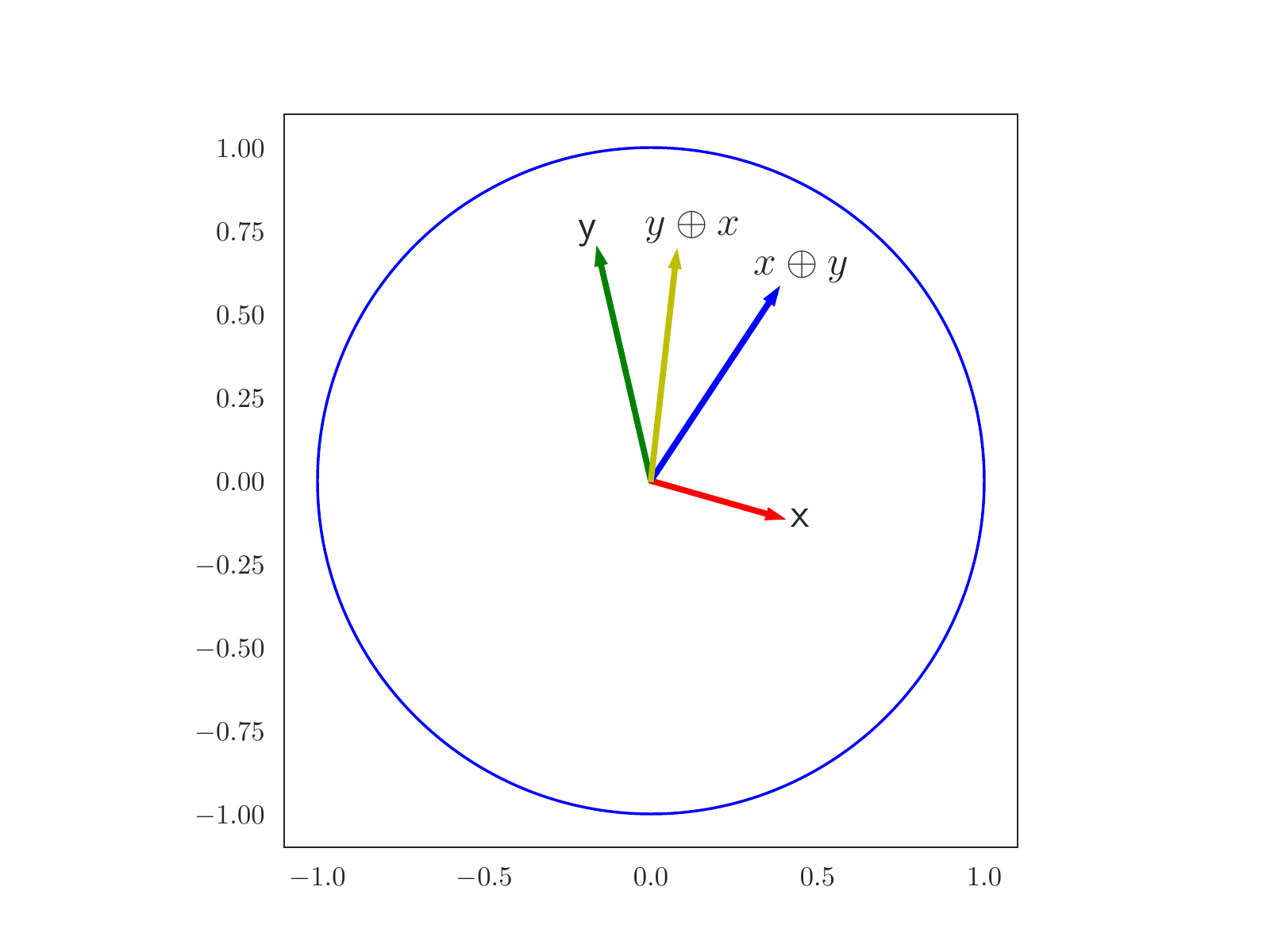}}}\hfill
\subfloat{\label{fig:mob-vec}{\includegraphics[clip, trim=2.5cm 0.8cm 3cm 1.2cm,width=.5\linewidth,keepaspectratio]{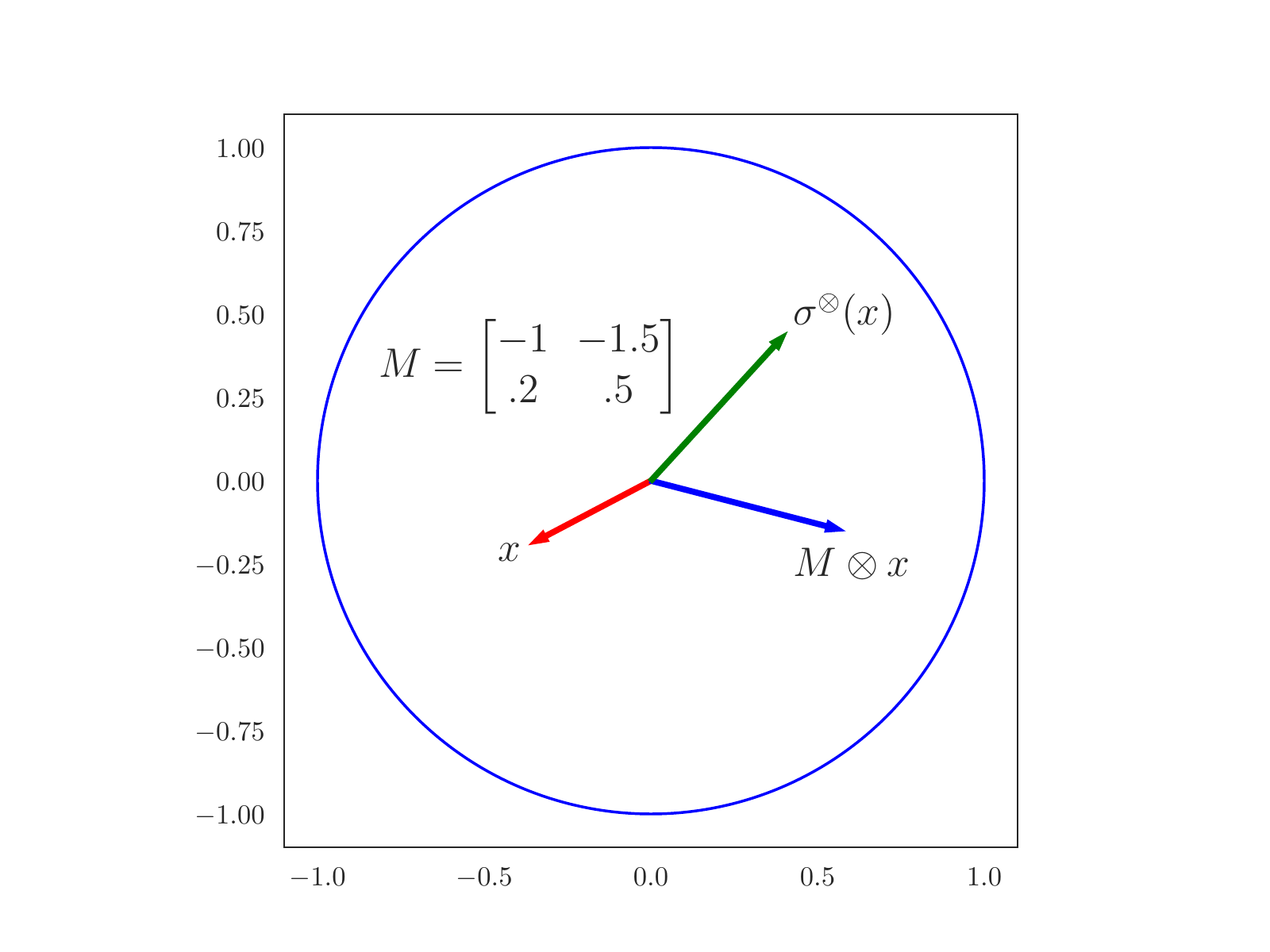}}}
\caption{Visualization of M\"obius operations. Left: M\"obius addition (noncommutative). Right: Matrix-vector multiplication and pointwise non-linearity.}
\label{fig:mob-op}
%\vspace{-3mm}
\end{figure}

\begin{figure*}[!t]
\centering
\includegraphics[height=4.5cm, keepaspectratio]{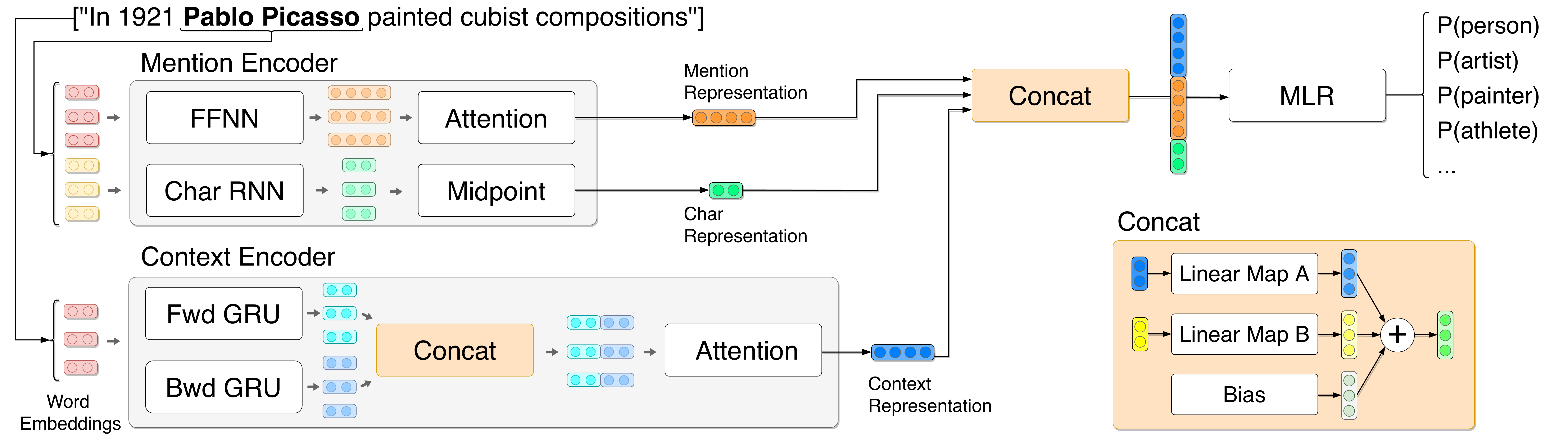}
%\vspace{-6mm}
\caption{Overview of the proposed model. The mention encoder extracts word and char-level entity representations. The context encoder is based on a bidirectional-GRU with attention. The outputs of both encoders are concatenated and passed to a classifier based on a multinomial logistic regression.}
\label{fig:model}
%\vspace{-2mm}
\end{figure*}

% Poincare ball 
\smallskip
\noindent
\textbf{Hyperbolic space:} It is a non-Euclidean space with constant negative curvature. %From the five isometric models for the hyperbolic space, 
We adopt the Poincar\'e ball model of hyperbolic space \cite{cannon1997hyperGeom}. In the general $n$-dimensional case, it becomes $\mathbb{D}^n = \{x \in \mathbb{R}^n \mid \|x\| < 1\}$\footnote{\newcite{ganea2018hyperNN} define the ball as $\mathbb{D}^n = \{x \in \mathbb{R}^n \mid c\|x\|^2 < 1\}$ with a parameter $c$ in relation to the radius of the Poincar\'e ball $r = 1 / \sqrt{c}$. In this work we assume $c = 1$ therefore we omit such parameter.}. The Poincar\'e model is a \textit{Riemannian manifold} equipped with the \textit{Riemannian metric} $g_{x}^{\mathbb{D}} = \lambda_{x}^{2} g^E$, where $\lambda_{x} := \frac{2}{1 - \|x\|^2}$
is called the \textit{conformal factor} and $g^{E} = \operatorname{I}_{n}$ is the Euclidean metric tensor. The distance between two points $x,y \in \mathbb{D}^n$ is given by:%
%\vspace{-4mm}
\begin{equation}
\small
d_{\mathbb{D}}(x, y) = \operatorname{cosh}^{-1}\left(1 + 2 \frac{\|x - y\|^2}{(1 - \|x\|^2)(1 - \|y\|^2)}\right)
%\vspace{-2mm}
\label{eq:hyper-dist}
\end{equation}%
% the model is also conformal to the euclidean model but I don't think that it is relevant to this paper.
% some intuition about the distance and the model...?
%Note that the hyperbolic distance between points grows exponentially as points move away from the center. This mirrors the exponential growth of the number of nodes in trees with increasing depths, thus making hyperbolic space a natural fit for representing trees and hence hierarchies \cite{krioukov2010hypernetworks,nickel2017poincare}.

\smallskip
\noindent
\textbf{M\"obius addition:} It is the hyperbolic analogous to vector addition in Euclidean space. Given two points $x, y \in \mathbb{D}^n$, it is defined as:
%\vspace{-2mm}
\begin{equation}
\small
%u \oplus v = \frac{(1 + 2c \langle x,y \rangle + c\|y\|^2)x + (1 - c\|x\|^2)y}{1 + 2c\langle x,y \rangle + c^2 \|x\|^2 \|y\|^2}    this one is with the c parameter
x \oplus y = \frac{(1 + 2 \langle x,y \rangle + \|y\|^2)x + (1 - \|x\|^2)y}{1 + 2\langle x,y \rangle + \|x\|^2 \|y\|^2}
%\vspace{-2mm}
\end{equation}
%Note that this operation is neither commutative nor associative.

\smallskip
\noindent
\textbf{M\"obius matrix-vector multiplication:} Given a linear map $M: \mathbb{R}^n \rightarrow \mathbb{R}^m$, which we identify with its matrix representation, and a point $x \in \mathbb{D}^n, Mx \ne 0$, it is defined as:
%\vspace{-2mm}
\begin{equation}
\small
    M \otimes x = \operatorname{tanh}\left(\frac{\|Mx\|}{\|x\|} \operatorname{tanh}^{-1}(\|x\|)\right) \frac{Mx}{\|Mx\|}
%\vspace{-2mm}
\end{equation}

\smallskip
\noindent
\textbf{Pointwise non-linearity:} If we model it as $\varphi: \mathbb{R}^n \rightarrow \mathbb{R}^n$, then its M\"obius version $\varphi^{\otimes}$ can be applied using the same formulation of the matrix-vector multiplication. A visualization of the aforementioned operations can be seen in Figure~\ref{fig:mob-op}. 

By combining these operations we obtain a one-layer feed-forward neural network (FFNN) in hyperbolic space, described as $y = \varphi^{\otimes} (M \otimes x \oplus b)$ with $M  \in \mathbb{R}^{m \times n}$ and $b \in \mathbb{D}^m$ as trainable parameters. %\footnote{More precisely, $M \in \mathcal{M}_{m,n}(\mathbb{R})$ where $\mathcal{M}$ is a \textit{Riemannian manifold}. Since for optimization we consider $M$ Euclidean, we omit the notation $\mathcal{M}$ throughout this work.}. 
Note that the parameter $b$ lies in the hyperbolic space, thus its updates during training need to be corrected for this geometry.

\smallskip
\noindent
\textbf{Exponential and logarithmic maps:} For each point $x \in \mathbb{D}^n$, let $T_x \mathbb{D}^n$ denote the associated \textit{tangent space}, which is always a subset of Euclidean space \cite{liu2019hypergraphsnn}. We make use of the exponential map $\operatorname{exp}_{x}: T_x \mathbb{D}^n \rightarrow \mathbb{D}^n$ and the logarithmic map $\operatorname{log}_{x}: \mathbb{D}^n \rightarrow T_x \mathbb{D}^n$ to map points in the hyperbolic space to the Euclidean space, and vice-versa.
%The mapping between the tangent space and hyperbolic space is done by . 
At the origin of the space, they are given for $v \in T_\textbf{0} \mathbb{D}^n \backslash \{0\}$ and $y \in \mathbb{D}^n \backslash \{0\}$: 
%\vspace{-2mm}
\begin{equation}
\small
\begin{aligned}
& \operatorname{exp}_\textbf{0}(v) = \operatorname{tanh}\left(\|v\| \right) \frac{v}{\|v\|}  \\
& \operatorname{log}_\textbf{0}(y) = \operatorname{arctanh}(\|y\|) \frac{y}{\|y\|}
\end{aligned}
%\vspace{-1mm}
\end{equation}

To map a point $y \in \mathbb{D}^n$ onto the Euclidean space we apply $\operatorname{log}_{\textbf{0}}(y)$. Conversely, to map a point $v \in \mathbb{R}^{n}$ onto the hyperbolic space, we assume $\mathbb{R}^{n} = T_{\textbf{0}} \mathbb{D}^n$ and apply $\operatorname{exp}_{\textbf{0}}(v)$. This allows to mix hyperbolic and Euclidean neural layers as shown in \textsection\ref{sec:system-ablation}. 
% More information about Riemannian geometry and M\"obius operations in Appendix~\ref{sec:appendix-riemmanian}.% and \ref{sec:appendix-mob-operations}.

%%%%%%%%%%%%%%%%%%%%%%%%%%%%%%%%%%%%%%%%%%%%%%%%%%%%%%%%%%%%%%%%%%%%%%%%
\section{Fine-grained Entity Typing}
\label{sec:task-definition}

Given a context sentence $s$ containing an entity mention $m$, the goal of entity typing is to predict the correct type labels $t_m$ that describe $m$ from a type inventory $T$. The ground-truth type set $t_m$ may contain multiple types, making the task a multi-class multi-label classification problem.

For fine-grained entity typing the type inventory $T$ tends to contain hundreds to thousands of labels. 
Encoding hierarchical information from large type inventories has been proven critical to improve performance \cite{lopez-etal-2019-fine}. 
Thus we hypothesize that our proposed hyperbolic model will benefit from this representation.

%%%%%%%%%%%%%%%%%%%%%%%%%%%%%%%%%%%%%%%%%%%%%%%%%%%%%%%%%%%%%%%%%%%%%%
\section{Hyperbolic Classification Model}
\label{sec:model}

In this section we propose a general hyperbolic neural model for classification with sequential data as input. The building blocks are defined in a generic manner such that they can be applied to different tasks, or integrated with regular Euclidean layers.
Our proposed architecture resembles recent neural models applied to entity typing \cite{choi2018ultra}. %It consists of word and char-level %entity mention encoder, and a context sentence 
%encoders based on recurrent neural networks with attention, followed by a classifier. 
% el bloque de arriba ya lo estoy diciendo aca abajo ahora, y con mas detalle.
For the encoders we employ the neural networks introduced in \newcite{ganea2018hyperNN}, we propose a novel attention mechanism operating entirely in the Poincar\'e model, and we extend the hyperbolic classifier to multi-class multi-label setups.
An overview of the model can be seen in Figure~\ref{fig:model}.

\subsection{Mention Encoder}
%\todo{cite Lee to justify architecture. Remove for space}
To represent the mention, we combine word and char-level features, similar to \newcite{lee2017e2ecoref}.
Given a sequence of $k$ tokens in a mention span, we represent them using pre-trained word embeddings $w_i \in \mathbb{D}^n$ which we assume to lie in hyperbolic space. 
%Since we intend to perform all operations under hyperbolic geometry, we use embeddings pre-trained in the Poincar\'e model. 
We apply a %matrix-vector multiplication over each word embedding, followed by a bias addition and a pointwise non-linearity. This is essentially a one-layer feed-forward neural network 
hyperbolic FFNN, described as:
%\vspace{-4mm}
\begin{equation}
m_i = \operatorname{tanh}^{\otimes} (W^M \otimes w_i \oplus b^M)
%\vspace{-1mm}
\end{equation}
with $m_i \in \mathbb{D}^{d_M}$, and where $W^M \in \mathbb{R}^{d_M \times n}, b^M \in \mathbb{D}^{d_M}$ are parameters of the model. %For $\varphi$ we use the hyperbolic tangent $\operatorname{tanh}$. % but any other non-linearity function could be applied. 
%some words in an entity mention may provide more useful information for typing therefore (next paragraph)
We combine the resulting $m_1, ..., m_k$ into a single mention representation $\mathbf{m} \in \mathbb{D}^{d_M}$ by computing a weighted sum of the token representations in hyperbolic space with the attention mechanism explained in \textsection\ref{sec:distance-attention}.

%To combine the results of the tokens into a single mention representation $\mathbf{M} \in \mathbb{D}^{d_M}$ we apply an attention mechanism (further explained in Section~\ref{sec:distance-attention}). This is essentially a weighted sum of the token representations, in the hyperbolic space.

Moreover, we extract features from the sequence of characters in the mention span with a recurrent neural network (RNN) \cite{lample2016neuralner}. We represent each character with a char-embedding $c_i \in \mathbb{D}^{d_C}$ that we train in the Poincar\'e ball. An RNN operating in hyperbolic space is defined by:
%\vspace{-2mm}
\begin{equation}
h_{t+1} = \varphi^{\otimes}(W^C \otimes h_t \oplus U^C \otimes c_t \oplus b^C)
%\vspace{-1mm}
\end{equation}
where $W^C, U^C \in \mathbb{R}^{d_C \times d_C}, b^C, h_t \in \mathbb{D}^{d_C}$, and $\varphi$ is a pointwise non-linearity function. Finally, we obtain a single representation $\mathbf{c} \in \mathbb{D}^{d_C}$ by taking the midpoint of the states $h_i$ using Equation~\ref{eq:midpoint}. %without the weights 
% \cite{ungar2010barycentric}.

%%%%%%%%%%%
\subsection{Context Encoder}
To encode the context we apply a hyperbolic version of gated recurrent units (GRU) \citep{cho2014gru} proposed in \newcite{ganea2018hyperNN}\footnote{For a complete description of this network see Appendix~\ref{sec:appendix-gru} or \newcite{ganea2018hyperNN} \textsection3.3}. Given a sequence of $l$ tokens, we represent them with a pre-trained word embedding $w_i \in \mathbb{D}^{n}$, and apply a forward and backward GRU, producing contextualized representations $\overrightarrow{h_i}, \overleftarrow{h_i} \in \mathbb{D}^{d_S}$ for each token. We concatenate the resulting states into a single embedding $s_i = \operatorname{concat}(\overrightarrow{h_i}, \overleftarrow{h_i})$ (see $\operatorname{concat}$ in \textsection\ref{sec:concat}), where $s_i \in \mathbb{D}^{2d_S}$. Ultimately, we combine $s_1,...,s_l$ into a single context representation $\mathbf{s} \in \mathbb{D}^{2d_S}$ with the distance-based attention mechanism.

%%%%%%%%%%%
\subsection{Concatenation}
\label{sec:concat}
If we model the concatenation of two vectors in the Poincar\'e ball as appending one to the other, this does not guarantee that the result remains inside the ball. Thus, we apply a generalized version of the concatenation operation. For $x \in \mathbb{D}^{k}, y \in \mathbb{D}^{l}$, then $\operatorname{concat}: \mathbb{D}^{k} \times \mathbb{D}^{l} \rightarrow \mathbb{D}^{n}$ is defined as:
%\vspace{-3mm}
\begin{equation}
\operatorname{concat}(x,y) = M_1 \otimes x \oplus M_2 \otimes y \oplus b
\label{eq:concat}
%\vspace{-3mm}
\end{equation}
where $M_1 \in \mathbb{R}^{n \times k}, M_2 \in \mathbb{R}^{n \times l}, b \in \mathbb{D}^{n}$ are parameters of the model. In Euclidean architectures, the concatenation of vectors is usually followed by a linear layer, which takes the form of Equation~\ref{eq:concat} when written explicitly.

%%%%%%%%%%%
\subsection{Distance-based Attention}
\label{sec:distance-attention}
Previous approaches to hyperbolic attention \cite{gulcehre2018hyperAttentionNet, chami2019hgcnn} require mappings of points to different spaces, which hinders their adoption into neural models. 
We propose a novel attention mechanism 
% Remove \cite{bahdanau2015attention, vaswani2017attention} cite for space
in the Poincar\'e model of hyperbolic space. We cast attention as a weighted sum of vectors in this geometry, without requiring any extra mapping of the inputs. In this manner, we make consistent use of the same analytical model of hyperbolic space across all components, which eases their integration.

%To generalize the mechanism of attention  in hyperbolic space, we propose . %\todo{removable}{Hence}, two questions arise: how to obtain the weights, and how to calculate a weighted sum in hyperbolic space. 
To obtain the attention weights, we %adapt the method proposed by \newcite{gulcehre2018hyperAttentionNet}, and 
exploit the hyperbolic distance between points \cite{gulcehre2018hyperAttentionNet}. 
% from https://arxiv.org/pdf/1909.12375.pdf Yi and Benjamin
% To encode subword order, "s" can be further enriched by a trainable position embedding p. We use addition to combine subword and position embeddings, namely s := s + p, which has become the de-facto standard method to encode positional information (Vaswani et al., 2017; Devlin et al., 2019)
Given a sequence of states $x_i \in \mathbb{D}^{n}$, we combine them with a trainable position embedding $p_i \in \mathbb{D}^{n}$ such that $r_i = x_i \oplus p_i$. We use addition as the standard method to enrich the states with positional information \cite{vaswani2017attention, devlin-etal-2019-bert}. We apply two different linear transformations on $r_i$ to obtain vectors $q_i$ and $k_i$, both lying in the Poincar\'e ball. We compute the distance between these two points and finally obtain the weight by applying a $\operatorname{softmax}$ over the sequence in the following manner:
%\vspace{-2mm}
\begin{equation}
\small
\begin{aligned}
q_i = W^Q \otimes r_i \oplus b^Q, \quad k_i = W^K \otimes r_i \oplus b^K \\
\alpha(q_i, k_i) = \operatorname{softmax}(-\beta d_{\mathbb{D}}(q_i, k_i))
\end{aligned}
\label{eq:attn-weights}
%\vspace{-2mm}
\end{equation}
where $W^Q, W^K \in \mathbb{R}^{n \times n}, b^Q, b^K \in \mathbb{D}^n$ and $\beta \in \mathbb{R}$ are parameters of the model. 
%Since distances in the Poincar\'e ball grow exponentially with the norm of the points, 
Attention weights will be higher for elements with $q_i$ and $k_i$ vectors placed close to each other.

The positional embeddings are trained along with the model as a hyperbolic parameter. For the context encoder, they reflect relative distances between the $i$-th word and the entity mention. For the mention encoder, they represent the absolute position of the word inside the mention span.

% justificacion del bias de https://arxiv.org/pdf/1606.02245.pdf: Second, we add a term a_q that allows to bias the attention mechanism towards words which tend to be important across the questions independently of the search key st−1. This is similar to what is achieved by the original attention mechanism proposed in (Bahdanau et al., 2015) without the burden of the additional tanh layer
% en caso que use la tanh y el bias, la attn se parece a esta: https://www.ijcai.org/proceedings/2017/0568.pdf

% una justificacion mas real es que ese bias hace que los vectores q y k se alejen del centro regardless del token que se esta procesando. Esto hace que la distancia no tienda a cero, capturando toda la atencion en un solo punto... ponele. Esto es:
% q = W^Q * h_i + b^Q
% k = W^K * h_i + b^K
% si h_i tiende a cero Y NO HAY BIAS, q y k tienden a cero y la distancia entre ellos tambien. Por lo tanto la attn va a ser mayor sobre el token de h_i. Voy a decir que NO quiero que pase eso y por esta razon agrego el bias en ambos cálculos.

To aggregate the points as a weighted summation in hyperbolic space we propose to apply the M\"obius midpoint, which obeys many of the properties that we expect from a weighted average in Euclidean space (\newcite{ungar2010barycentric}, Theorem 4.6):
%\vspace{-2mm}
\begin{equation}
m = \frac{1}{2} \otimes \frac{\sum_{i=1}^n \alpha_i \gamma(x_i)^2 x_i}{\sum_{i=1}^n \alpha_i \left( \gamma(x_i)^2 - \frac{1}{2} \right)}
\label{eq:midpoint}    
%\vspace{-2mm}
\end{equation}
where $x_i$ are the states in the sequence, $\alpha_i$ the weights corresponding to each state, and $\gamma(x_i)$ the Lorentz factors. By applying the M\"obius midpoint 
%instead of the Einstein midpoint as in \citet{gulcehre2018hyperAttentionNet}, 
we develop an attention mechanism that operates entirely in the Poincar\'e model of hyperbolic space. Detailed formulas and experimental observations can be found in Appendix~\ref{sec:appendix-attention}.

\subsection{Classification in the Poincar\'e Ball}
\label{sec:mlr}

The input of the classifier is the concatenation of mention and context features.
%$\operatorname{concat}(\mathbf{m}, \mathbf{c}, \mathbf{s}) \in \mathbb{D}^m$ with $m = d_M + d_C + 2 d_S$.
To perform multi-class classification in the Poincar\'e ball, we adapt the generalized multinomial logistic regression (MLR) from \newcite{ganea2018hyperNN}. 
Given $K$ classes and $k \in \{1,...,K\}$, $p_k \in \mathbb{D}^m$, $a_k \in T_{p_k} \mathbb{D}^m \backslash \{0\}$, the formula for the hyperbolic MLR is:
%\vspace{-2mm}
\begin{equation}
\resizebox{.85\hsize}{!}{$
\begin{multlined}
p(y=k|x) \propto \\
f\left(\lambda_{p_k} \|a_k\| \operatorname{sinh}^{-1} \left(\frac{2 \langle -p_k \oplus x, a_k\rangle}{(1 - \| -p_k \oplus x \|^2)\|a_k\|} \right) \right)
\end{multlined}
$}
\label{eq:mlr}
%\vspace{-2mm}
\end{equation}

Where $x \in \mathbb{D}^m$, and $p_k$ and $a_k$ are trainable parameters. %Since $a_k \in T_{p_k} \mathbb{D}^m$ and it depends on $p_k$, it is replaced by $a_k = (\lambda_0 / \lambda_{p_k})a_{k}^{\prime}$, where $a_{k}^{\prime} \in T_{\textbf{0}} \mathbb{D}^m = \mathbb{R}^m$. 
It is based on formulating logits as distances to margin hyperplanes. %\todo{Removable}{In Euclidean space, hyperplanes can be specified with a point of origin and a normal vector.} 
The hyperplanes in hyperbolic space %for a point $p \in \mathbb{D}^m$ and $a \in T_{p} \mathbb{D}^m \backslash \{0\}$ 
are defined by the union of all geodesics passing through $p_k$ and orthogonal to $a_k$.
%In the case of multi-class one-label classification, $f$ is replaced by $\operatorname{exp}$ with the appropriate normalization to obtain a $\operatorname{softmax}$.

Although this formulation was made for one-label classification, the underlying notion also holds for multi-label setups. In that case, we need to be able to select several classes by considering the distances (logits) to all hyperplanes. To achieve that we employ the $\operatorname{sigmoid}$ function as $f$, instead of a $\operatorname{softmax}$, and predict the given class if $p(y=k|x) > 0.5$. More details in Appendix~\ref{sec:appendix-mlr}.

Figure~\ref{fig:hyperplanes} shows examples of the hyperbolic definition of multiple hyperplanes, which follow the curvature of the space.

\subsection{Optimization}
With the proposed classification model, we aim to minimize variants of the binary cross-entropy loss function as the training objective.

The model has trainable parameters in both Euclidean and hyperbolic space. We apply the Geoopt implementation of \textit{Riemannian Adam} \cite{geoopt2019kochurov} as a Riemannian adaptive optimization method \cite{becigneul2019riemannianMethods} to carry out a gradient-based update of the parameters in their respective geometry.

\begin{table*}[!th]
\small
\centering
\adjustbox{max width=\textwidth}{
\begin{tabular}{lccccccccccccr}
\toprule
\textsc{} & \multicolumn{3}{c}{\textbf{Total}} & \multicolumn{3}{c}{\textbf{Coarse}} & \multicolumn{3}{c}{\textbf{Fine}} & \multicolumn{3}{c}{\textbf{Ultra-Fine}} & \multicolumn{1}{l}{\textbf{}} \\ \cmidrule(lr){2-4}\cmidrule(lr){5-7}\cmidrule(lr){8-10}\cmidrule(lr){11-13}
\textbf{Model} & \textbf{P} & \textbf{R} & \textbf{\texorpdfstring{F\textsubscript{1}}} & \textbf{P} & \textbf{R} & \textbf{\texorpdfstring{F\textsubscript{1}}} & \textbf{P} & \textbf{R} & \textbf{\texorpdfstring{F\textsubscript{1}}} & \textbf{P} & \textbf{R} & \textbf{\texorpdfstring{F\textsubscript{1}}} & \textbf{\# Params} \\
\hline
\textsc{Denoised} & 50.7 & 33.1 & \textbf{40.1} & 66.9 & \textbf{80.7} & 73.2 & 41.7 & 46.2 & 43.8 & 45.6 & 17.4 & 25.2 & 31.0M \\
\textsc{BERT} & \textbf{51.6} & 32.8 & \textbf{40.1} & \textbf{67.4} & 80.6 & \textbf{73.4} & 41.6 & \textbf{54.7} & \textbf{47.3} & \textbf{46.3} & 15.6 & 23.4 & 110.0M \\
\textsc{LabelGCN} & 49.3 & 28.1 & 35.8 & 66.2 & 68.8 & 67.5 & \textbf{43.9} & 40.7 & 42.2 & 42.4 & 14.2 & 21.3 & 5.1M \\
\textsc{MultiTask} & 48.0 & 23.0 & 31.0 & 60.0 & 61.0 & 61.0 & 40.0 & 38.0 & 39.0 & 42.0 & 8.0 & 14.0 & 6.1M \\
\hline
\textsc{hy} \modbase & 48.5 & 29.1 & 36.3 & 64.4 & 72.2 & 68.1 & 39.4 & 38.5 & 38.9 & 39.3 & 14.5 & 21.2 & 1.8M \\
\textsc{hy} \modlarge & 42.3 & 33.5 & 37.4 & 63.6 & 72.1 & 67.6 & 36.3 & 48.3 & 41.4 & 33.3 & 19.7 & 24.7 & 4.6M \\
\textsc{hy} \modxlarge & 43.4 & \textbf{34.2} & 38.2 & 61.4 & 73.9 & 67.1 & 35.7 & 46.6 & 40.4 & 36.5 & \textbf{19.9} & \textbf{25.7} & 9.5M \\
\specialrule{.1em}{.05em}{.05em}
\end{tabular}
}
%\vspace{-1mm}
\caption{Macro-averaged P, R and $\operatorname{F}_1$ on the Ultra-Fine dev set for different baselines and models. We only reproduced \textsc{LabelGCN}. Values for other baselines are taken from the original publications.}
\label{tab:all-results}
%\vspace{-3mm}
\end{table*}

%%%%%%%%%%%%%%%%%%%%%%%%%%%%%%%%%%%%%%%%%%%%%%%%%%%%%%%%%%%%%%%%%%%%%%
\section{Experiments}
We evaluate the proposed hyperbolic model on two different datasets for fine-grained entity typing, and compare to Euclidean baselines as well as state-of-the-art models.

%Thus we hypothesize that our proposed hyperbolic model will benefit from this representation.

%A natural way to deal with a large number of types is to organize them in a hierarchy ranging from general, \textit{coarse} types near the top, to more specific, \textit{fine} types at the bottom \cite{lopez-etal-2019-fine}. 
%Since hyperbolic geometry is naturally equipped to model hierarchical structures, 

%%%%%%%%%%%
\subsection{Data}
For analysis and evaluation of the model, we focus on the Ultra-Fine entity typing dataset \cite{choi2018ultra}. %introduced in \newcite{choi2018ultra}. 
It contains 10,331 target types defined as free-form noun phrases and divided in three levels of granularity: \textit{coarse}, \textit{fine} and \textit{ultra-fine}. Besides this segregation, the dataset does not provide any further explicit information about the relations among the types. The data consist of 6,000 crowdsourced examples and 6M training samples in the open-source version, automatically extracted with distant supervision. %, by entity linking and nominal head word extraction. 
Our evaluation is done on the original crowdsourced dev/test splits.

To gain a better understanding of the proposed model, we also experiment on the OntoNotes dataset \cite{gillick2014context} as it is a standard benchmark for entity typing.

\subsection{Setup}

%We regard the dimensions $d_M, d_C, d_S$ as hyperparameters. 
The MLR classifier operates in a hyperbolic space of $m$ dimensions with $m = d_M + d_C + 2 d_S$. By setting different values, we experiment with three models: \modbase\ $(m = 100)$, \modlarge\ $(m = 250)$ and \modxlarge\ $(m = 500)$.

As word embeddings we employ Poincar\'e GloVe embeddings \cite{tifrea2018poincareGlove}, which are pre-trained in the Poincar\'e model. Hence, the input to the encoders is already in hyperbolic space and all operations can be performed in this geometry. These embeddings are not updated during training.
Low values of dropout are used since the model was very sensitive to this parameter given the behaviour of the hyperbolic distance. % with respect to the norm of the points. 
%\todo{removable}{Our} model is implemented in PyTorch \cite{paszke2017pytorch}.

On the Ultra-Fine dataset, for each epoch, we train over the entire training set, and we run extra iterations over the crowdsourced split before evaluating. In this way, the model benefits from the large amount of noisy, automatically-generated data, and is fine-tuned with high-quality crowdsourced samples. As previous work \cite{xiong2019inductiveBias, onoe-durrett-2019-denoise}, we optimize the multi-task objective proposed by \newcite{choi2018ultra}.

For evaluation we report Macro-averaged and Micro-averaged $\operatorname{F}_1$ metrics computed from the precision/recall scores over the same three granularities established by \newcite{choi2018ultra}.
For all models we optimize \textit{Total} Macro-averaged $\operatorname{F}_1$ on the validation set, and evaluate on the test set. Following \newcite{ganea2018hyperNN}, we report the average of three runs given the highly non-convex spectrum of hyperbolic neural networks. Hyperparameters are detailed in Appendix~\ref{sec:appendix-hyperparams} along with other practical aspects to ensure numerical stability.

%%%%%%%%%%%%%%%%%%%%%%%%%%%%%%%%%%
\subsection{Baselines}
\noindent
\textbf{Euclidean baseline:} We replace all operations of the hyperbolic model by their Euclidean counterpart. % and apply the same training setup as in the hyperbolic case.
To map the Poincar\'e GloVe embeddings to the Euclidean space we apply $\operatorname{log}_{\textbf{0}}$. %the logarithmic map at $x = 0$.
We do not apply any kind of normalization or correction over the weights to circumscribe them into the unit ball. On the contrary, we grant them freedom over the entire Euclidean space to establish a fair comparison.

\noindent
\textbf{Multi-task:} Model proposed by \newcite{choi2018ultra}, along with the Ultra-Fine dataset.

\noindent
\textbf{LabelGCN:} Model introduced by \newcite{xiong2019inductiveBias}. A label-relational inductive bias is imposed by means of a graph propagation layer that encodes label co-occurrence statistics.

\noindent
\textbf{BERT:} We compare to the setup of \newcite{onoe-durrett-2019-denoise} in which BERT \cite{devlin-etal-2019-bert} is adapted for this task and  fine-tuned on the crowdsourced train split.

\noindent
\textbf{Denoised:} An ELMo-based model \cite{peters2018elmo} proposed by \newcite{onoe-durrett-2019-denoise} trained on raw and denoised distantly-labeled data.

%%%%%%%%%%%%%%%%%%%%%%%%%%%%%%%%%%%%%%%%%%%%%%%%%%%%%%%%%%%%%%%%%%%%
\section{Results and Discussion}
\label{sec:results-and-discussion}

\begin{figure}[!b]
%\vspace{-4mm}
\small
\centering
\subfloat[Euclidean Space.]{\label{fig:hyperplanes-eu}{\includegraphics[width=.47\linewidth,keepaspectratio]{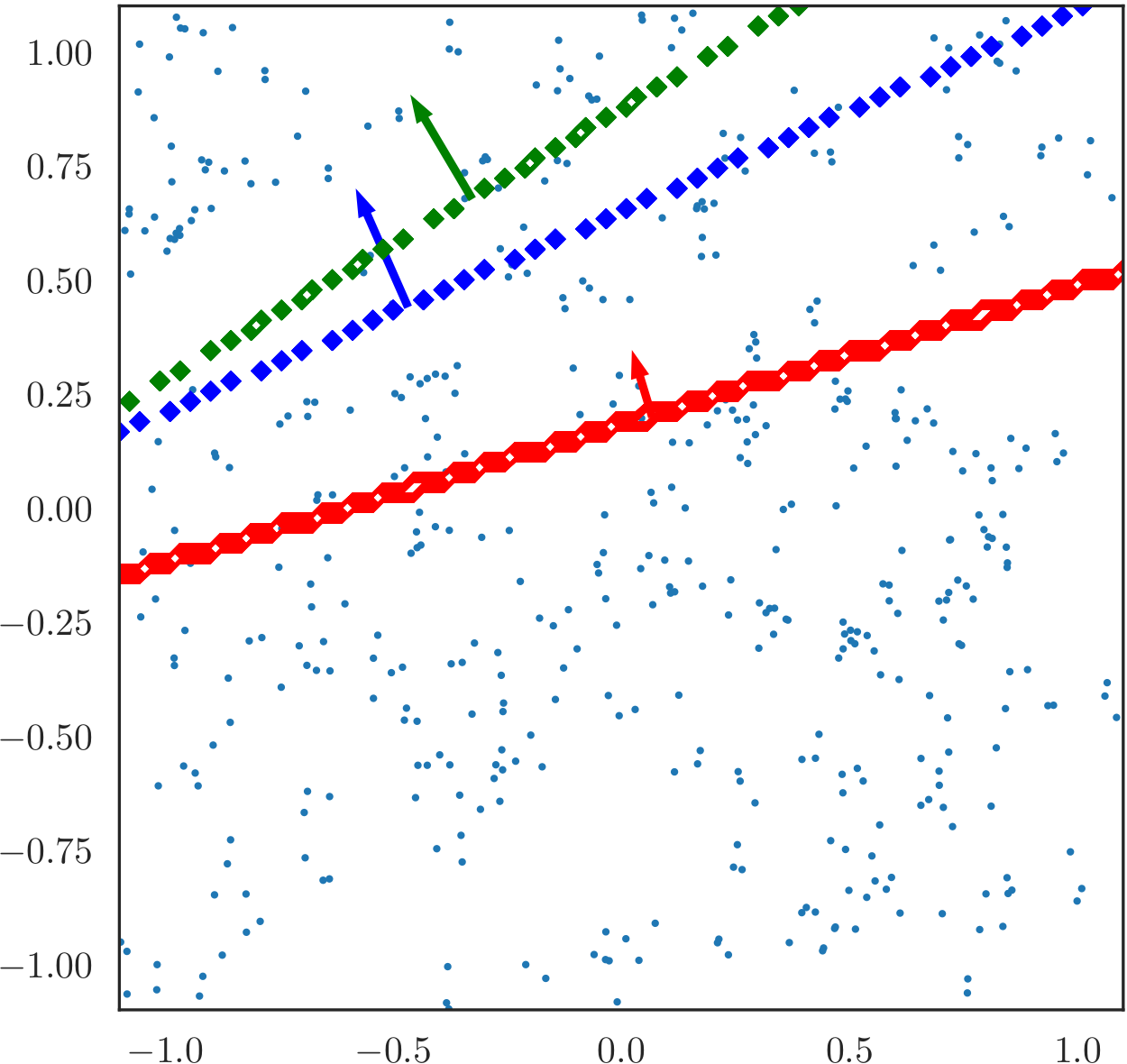}}}\hfill
\subfloat[Hyperbolic Space.]{\label{fig:hyperplanes-hy}{\includegraphics[width=.47\linewidth,keepaspectratio]{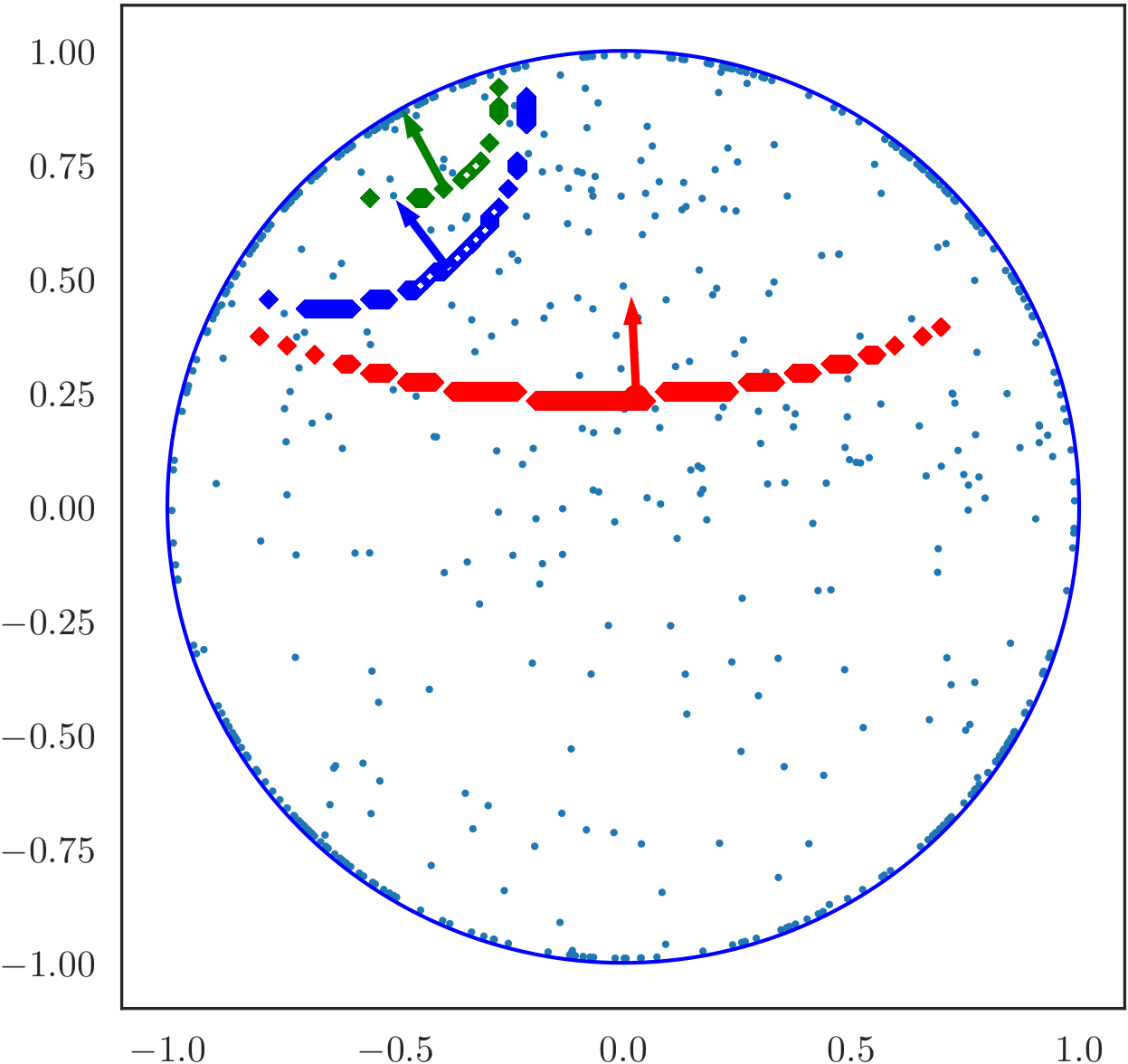}}}
%\vspace{-2mm}
\caption{Classification hyperplanes for the types \texttt{person} (red), \texttt{artist} (blue) and \texttt{musician} (green). The hyperbolic formulation of the hyperplanes is better suited for hierarchical inventories.}
\label{fig:hyperplanes}
%\vspace{-2mm}
\end{figure}

Following previous work \cite{choi2018ultra, onoe-durrett-2019-denoise}, we report results on the development set in Table~\ref{tab:all-results}. All hyperbolic models outperform \textsc{MultiTask} and \textsc{LabelGCN} baselines on \textit{Total} Macro $\operatorname{F}_1$. 
% reconocer la derrota
\textsc{Denoised} and \textsc{BERT} systems, based on large pre-trained models, show the best \textit{Total} performance. 
% sos bueno en todo, pero les ganas en ultra-fine grained. Por que?
Nonetheless, \textsc{hy xLarge} has a competitive performance, and surpasses both systems on \textit{ultra-fine} \fone. 
In the hyperbolic model, fine-grained types are placed near the boundary of the ball, where the amount of space grows exponentially. Furthermore, the underlying structure of the type inventory is hierarchical, thus the hyperbolic definition of the hyperplanes is well-suited to improve the classification in this case (see comparison with Euclidean classifiers on Figure~\ref{fig:hyperplanes}). These properties combined enable the hyperbolic model to excel at classifying hierarchical labels, with outstanding improvements on very fine-grained types.

% lo logras con muchos menos parametros. Por que?
The reduction of the parameter size is also remarkable: $70\%$ and $91\%$ versus \textsc{Denoised} and \textsc{BERT} respectively. 
This emphasizes the importance of choosing a suitable metric space that fits the data distribution (hierarchical in this case) as a powerful and efficient inductive bias. Through adequate tools and formulations, we are able to exploit this bias without introducing an overload on the parameter cost.

Correspondence of results between \textsc{hy base} and \textsc{LabelGCN} suggest that both models capture similar information. \textsc{LabelGCN} requires label co-occurrence statistics represented as a weighted graph, from where a hierarchy can be easily derived \cite{krioukov2010hypernetworks}. The similarity of results indicates that the hyperbolic model is able to implicitly  encode the latent hierarchical information in the label co-occurrences without additional inputs or the burden of the graph layer. 

%\todo[inline]{OR: The inductive bias is given by the geometrical properties of the hyperbolic representation rather than explicitly imposed by the data encoded on the graph.}

\begin{table}[!t]
%\vspace{-4mm}
\small
\centering
\adjustbox{max width=\linewidth}{
\begin{tabular}{lclclc}
\toprule
\multicolumn{2}{c}{\texttt{person}} & \multicolumn{2}{c}{\texttt{artist}} & \multicolumn{2}{c}{\texttt{musician}} \\
\textbf{Types} & $d_{\mathbb{D}}$ & \textbf{Types} & $d_{\mathbb{D}}$ & \textbf{Types} & $d_{\mathbb{D}}$ \\
\midrule
\textit{artist} & 0.26 & \textit{musician} & 0.25 & \textit{singer} & 0.24 \\
author & 0.28 & actor & 0.26 & actor & 0.25 \\
actor & 0.30 & person & 0.26 & artist & 0.25 \\
speaker & 0.30 & author & 0.26 & composer & 0.27 \\
leader & 0.30 & singer & 0.28 & band & 0.27 \\
\bottomrule
\end{tabular}
}
%\vspace{-2mm}
\caption{Closest $p_k$ points in the Poincar\'e Ball to different Ultra-Fine entity types. The model is able to  capture hierarchical relations such as \texttt{singer} \textit{is-a} \texttt{musician} \textit{is-a} \texttt{artist} \textit{is-a} \texttt{person}.}
\label{tab:closest-pk-hier}
%\vspace{-2mm}
\end{table}

%HERE: En este parrafo, destacar que a pesar de que la jerarquía esta implicita, el modelo la aprende automaticamente!!!!!
To shed light on this aspect, we inspect the points $p_k$ learned by \textsc{hy base} to define the hyperplanes of Equation~\ref{eq:mlr}. Table~\ref{tab:closest-pk-hier} shows the types corresponding to the closest points to the label \texttt{person} and its \textit{subtypes}, measured by hyperbolic distance. The types are highly correlated given that they often co-occur in similar contexts. Moreover, the model captures hyponymic relations (\textit{is-a}) present in the label co-occurrences. An analogous behaviour is observed for other types (see tables in Appendix~\ref{sec:appendix-closest-pk}). The inductive bias given by the hyperbolic geometry allows the model to capture the hierarchy, deriving a meaningful and interpretable representation of the label space: \textit{coarse} labels near the origin, \textit{fine-grained} labels near the boundary, and hyponymic relations are preserved. It is also noteworthy that the model learns these relations automatically without requiring the explicit data encoded in the graph.

%%%%%%%%%%%%%%%%%%%%%%%%%%%%%%%
\subsection{Comparison of the Spaces}

A comparison of the metric spaces for different models on the test set is shown in Table~\ref{tab:results-hy-vs-eu}. It can be seen that the hyperbolic model outperforms its Euclidean variants in all settings.
It is notable that this trend holds even in high-dimensional spaces ($500$ dimensions for \modxlarge). Since the label inventory exhibits a clearly hierarchical structure, it perfectly suits the hyperbolic classification method.

The hyperbolic model brings considerable gains as the granularity becomes finer: $5.1\%$ and $16.2\%$ relative improvement in \textit{fine} and \textit{ultra-fine} Macro $\operatorname{F}_1$ respectively for the \modbase\ model over its Euclidean counterpart. 
We also observe that as the size of the model increases, the Euclidean baseline becomes more competitive for \textit{ultra-fine}. This is due to the Euclidean model gaining enough capacity to accommodate the separation hyperplanes with higher dimensions, thus reducing the gap.

It is noticeable that the \modbase\ model outperforms the larger ones on \textit{coarse} and \textit{fine} granularities. That is due to the larger models overfitting %on the train set 
given the low dropout applied. 
Moreover, Euclidean and hyperbolic models exhibit a similar performance on the \textit{coarse} granularity when compared to each other.
%This is, Euclidean hyperplanes are able to classify \textit{coarse} labels as well as hyperbolic hyperplanes.
A possible explanation is that the separation planes for these labels are located closer to the origin of the space.
In this region, the spaces behave alike in terms of the distance calculation, and this similarity is reflected in the results as well. 
%\todo{If I cmp dims vs skipgram or ganea this is the place} 
%Further analysis in Section~\ref{sec:system-ablation}.
%En caso de que lo haga, por aca seria el lugar para decir que skipgram o poincare embeds no comparan con grandes dims (ver en mix-curvature que los nombran) y que en nuestro caso, aun usando 500 dimensiones para clasificar, el modelo hyperbolico le sigue ganando al euclideo

\begin{table}[!t]
\centering
\small
\adjustbox{max width=\linewidth}{
\begin{tabular}{p{1cm}ccccccc}
\toprule
\multicolumn{2}{l}{\textsc{}} &  \multicolumn{2}{c}{\textbf{Coarse}} & \multicolumn{2}{c}{\textbf{Fine}} & \multicolumn{2}{c}{\textbf{Ultra}} \\
\multicolumn{1}{c}{\textbf{Model}} &  & \textbf{Ma} & \textbf{Mi} & \textbf{Ma} & \textbf{Mi} & \textbf{Ma} & \textbf{Mi} \\
\hline
\multirow{2}{*}{\modbase} & \textsc{hy} & \textbf{69.6} & \textbf{67.3} & \textbf{42.0} & \textbf{39.7} & 21.2 & 19.1 \\
 & \textsc{eu} & 68.5 & 66.1 & 39.8 & 36.5 & 17.8 & 16.1 \\
 \hline
\multirow{2}{*}{\modlarge} & \textsc{hy} & 67.9 & 65.4 & 38.4 & 36.3 & 24.3 & 22.3 \\
 & \textsc{eu} & 67.1 & 63.8 & 36.7 & 34.7 & 22.0 & 19.7 \\
 \hline
\multirow{2}{*}{\modxlarge} & \textsc{hy} & 69.1 & 66.2 & 39.7 & 37.2 & \textbf{26.1} & \textbf{24.0} \\
 & \textsc{eu} & 67.9 & 65.4 & 37.8 & 35.3 & 22.2 & 20.0 \\
\specialrule{.1em}{.05em}{.05em}
\end{tabular}
}
%\vspace{-2mm}
\caption{Results on Ultra-Fine test set for macro and micro \fone{} across metric spaces and dimensions.}
\label{tab:results-hy-vs-eu}
%\vspace{-4mm}
\end{table}

\begin{table}[!b]
\small
\centering
\adjustbox{max width=\linewidth}{
\begin{tabular}{lcccccc}
\toprule
\textsc{base} & \multicolumn{2}{c}{\textbf{Coarse}} & \multicolumn{2}{c}{\textbf{Fine}} & \multicolumn{2}{c}{\textbf{Ultra}} \\
\textbf{Model} & \textbf{Ma} & \textbf{Mi} & \textbf{Ma} & \textbf{Mi} & \textbf{Ma} & \textbf{Mi} \\
\midrule
\textsc{hy GloVe} & \textbf{68.7} & \textbf{66.6} & \textbf{41.5} & \textbf{38.8} & \textbf{22.1} & \textbf{20.1} \\
\textsc{eu GloVe} & 67.8 & 65.3 & 39.7 & 36.0 & 20.7 & 18.6 \\
\bottomrule
\end{tabular}
}
%\vspace{-2mm}
\caption{Test results on Ultra-Fine. Poincar\'e GloVe embeddings %\cite{tifrea2018poincareGlove} 
are replaced by regular GloVe embeddings. %\cite{pennington2014glove}.
}
\label{tab:word-embedding-ablation}
%\vspace{-4mm}
\end{table}

\subsubsection{Word Embeddings Ablation}
\label{sec:embed-ablation}

The input for both the Euclidean and hyperbolic models are Poincar\'e GloVe embeddings, which are originally trained in hyperbolic space \cite{tifrea2018poincareGlove}. This might favor the hyperbolic model, despite the application of the $\operatorname{log}_{\textbf{0}}$ map in the Euclidean case. Thus, we replace the hyperbolic embeddings by the regular GloVe embeddings \cite{pennington2014glove}, and use $\operatorname{exp}_{\textbf{0}}$ on the hyperbolic model to project them into the ball. 

Table~\ref{tab:word-embedding-ablation} shows that the tendency of the \modbase\ hyperbolic model outperforming the Euclidean one holds, and that the improvement does not come from the embeddings. Also, in this way we showcase how the hyperbolic model can be easily integrated with regular word embeddings.

%%%%%%%%%%%%%%%%%%%%%%%%%%%%
\subsubsection{Component Ablation}
\label{sec:system-ablation}

\begin{table}[!t]
%\vspace{-3mm}
\small
\centering
\adjustbox{max width=\linewidth}{
\begin{tabular}{lcccccc}
\toprule
\textsc{} & \multicolumn{2}{c}{\textbf{Coarse}} & \multicolumn{2}{c}{\textbf{Fine}} & \multicolumn{2}{c}{\textbf{Ultra}} \\
\textbf{Model} & \textbf{Ma} & \textbf{Mi} & \textbf{Ma} & \textbf{Mi} & \textbf{Ma} & \textbf{Mi} \\
\hline
\textsc{hy base} & \textbf{69.6} & \textbf{67.3} & \textbf{42.0} & \textbf{39.7} & 21.2 & 19.1 \\
\hline
\textsc{Eu} Encoder & 68.8 & 66.3 & 41.7 & 38.9 & \textbf{22.0} & \textbf{20.1} \\
\textsc{Eu} Attention & 68.9 & 66.4 & 40.8 & 38.0 & 20.1 & 18.4 \\
\textsc{Eu} Concat & 68.6 & 66.1 & 40.6 & 37.5 & 21.8 & 19.8 \\
\textsc{Eu} MLR & 69.2 & 67.1 & 40.8 & 38.0 & 17.3 & 15.8 \\
\specialrule{.1em}{.05em}{.05em}
\end{tabular}
}
%\vspace{-1mm}
\caption{Results on Ultra-Fine test set. Ablation of the hyperbolic model, replacing one component by its Euclidean counterpart at a time.}
\label{tab:component-ablation}
%\vspace{-2mm}
\end{table}

With the aim of analyzing the contribution of the different hyperbolic components, we perform an ablation study on the \modbase\ model. We divide the system in \textit{encoder}, \textit{attention} (both in the mention and context encoders), \textit{concatenation}, and \textit{MLR}, and replace them, one at a time, by their Euclidean counterparts. Note that when Euclidean and hyperbolic components are mixed, we convert the internal representations from one manifold to the other with the $\operatorname{exp}_{\textbf{0}}$ and $\operatorname{log}_{\textbf{0}}$ maps.

As we see in Table~\ref{tab:component-ablation}, MLR is the component that contributes the most to the \textit{ultra-fine} classification.
The hierarchical structure of the type inventory combined with the hyperbolic definition of the hyperplanes are the reason of this (see Figure~\ref{fig:hyperplanes}).

Hyperbolic attention and concatenation are relevant for \textit{coarse} and \textit{fine-grained} classification (here is where the biggest drop appears when they are removed), but do not play a major role in the \textit{ultra-fine} granularity. 

Finally, the encoders do not benefit from the hyperbolic representation. As the reason for this we consider that the model is not able to capture tree-like relations among the input tokens such that they can be exploited for the task.

This ablation suggests that the main benefits of hyperbolic layers arise when they are incorporated at deeper levels of representation in the model, and not over low-level features or raw text. %\todo{Removable}{Furthermore}, when the underlying class distribution is hierarchical, classification in hyperbolic space brings considerable gains over Euclidean approaches.

\smallskip
\noindent
\textbf{Computing time:} M\"obius operations are more expensive than their Euclidean counterparts. Due to this, in our experiments we found the hyperbolic encoder to be twice slower, and the MLR $1.5$ times slower than their Euclidean versions.

\begin{table}[!t]
\centering
\small
\adjustbox{max width=0.97\linewidth}{
\begin{tabular}{cccccccc}
\specialrule{.1em}{.05em}{.05em}
\multicolumn{2}{l}{\textsc{}} &  \multicolumn{2}{c}{\textbf{Coarse}} & \multicolumn{2}{c}{\textbf{Fine}} & \multicolumn{2}{c}{\textbf{Ultra}} \\
\textbf{Model} & \textbf{} & \textbf{Ma} & \textbf{Mi} & \textbf{Ma} & \textbf{Mi} & \textbf{Ma} & \textbf{Mi} \\
\hline
\multirow{2}{*}{\textsc{base}} & \textsc{hy} & 82.0 & 80.2 & 41.8 & \textbf{41.4} & 23.9 & 25.0 \\
 & \textsc{eu} & 81.8 & 80.3 & 37.7 & 36.1 & 17.5 & 15.8 \\
 \hline
\multirow{2}{*}{\textsc{large}} & \textsc{hy} & \textbf{83.1} & \textbf{81.3} & \textbf{42.0} & \textbf{41.4} & \textbf{24.0} & \textbf{25.2} \\
 & \textsc{eu} & 82.4 & 80.9 & 38.2 & 36.7 & 18.9 & 18.1 \\
 \specialrule{.1em}{.05em}{.05em}
\end{tabular}
}
%\vspace{-2mm}
\caption{Macro and micro \fone\ on OntoNotes test set for different granularities.}
\label{tab:onto-granular}
%\vspace{-1mm}
\end{table}

\begin{table}[!b]
%\vspace{-2mm}
\centering
\small
\begin{tabular}{lccc}
\specialrule{.1em}{.05em}{.05em}
\textbf{Model} & \textbf{Acc.} & \textbf{Ma-\texorpdfstring{F\textsubscript{1}}} & \textbf{Mi-\texorpdfstring{F\textsubscript{1}}} \\
\hline
\newcite{shimaoka2017neural} & 51.7 & 70.9 & 64.9 \\
\textsc{AFET} \cite{ren2016AFET} & 55.1 & 71.1 & 64.7 \\
\textsc{PLE} \cite{ren2016LabelNoiseReduction} & 57.2 & 71.5 & 66.1 \\
\textsc{BERT}  & 51.8 & 76.6 & 69.1 \\
\textsc{MultiTask} & 59.5 & 76.8 & 71.8 \\
\textsc{LabelGCN} & \textbf{59.6} & \textbf{77.8} & \textbf{72.2} \\
\hline
\textsc{hy large} & 47.4 & 75.8 & 69.4 \\
\specialrule{.1em}{.05em}{.05em}
\end{tabular}
%\vspace{-2mm}
\caption{Total accuracy, macro and micro \fone\ scores on OntoNotes test set.}
\label{tab:onto-cmp}
%\vspace{-4mm}
\end{table}

%%%%%%%%%%%%%%%%%%%%%%%%%%%%%%%%%%%%
\subsection{OntoNotes Dataset}

To further understand the capabilities of the proposed model we also perform an evaluation on the OntoNotes dataset \cite{gillick2014context}. In this case, we apply the standard binary cross-entropy loss, since %the MultiTask objective retrieves an excessive amount of 
fine-grained labels are scarce in this dataset. Following previous work \cite{xiong2019inductiveBias}, we train over the dataset augmented by \newcite{choi2018ultra}. Results for the three granularities for \modbase\ and \modlarge\ models are presented in Table~\ref{tab:onto-granular}. The hyperbolic models outperform the Euclidean baselines in both cases, and the difference is noticeable for \textit{fine} and \textit{ultra-fine} ($42.0$ vs $38.2$ and $24.0$ vs $18.9$ on Macro \fone\ for the \modlarge\ model), in accordance with the results on Ultra-Fine. 

We report a comparison with neural systems in Table~\ref{tab:onto-cmp}. The hyperbolic model, without requiring the explicit hierarchy provided in this dataset, achieves a competitive performance. Nonetheless, the advantages of the hyperbolic model are mitigated by the low multiplicity of fine-grained labels, and the lower hierarchy.

%%%%%%%%%%%%%%%%%%%%%%%%%%%%%%%%%%%%%%%%%%%%%%%%%%%%%%%%%%%%
\section{Related Work}
\label{sec:related-work}

% Podemos hablar de:
% - Hyperbolic spaces que entran mas bien al final y no a lo largo de la representacion: Lopez o Tay (HyperQA).
% - Their perfomance has been showcased over synthetic datasets (dependiendo de si bardie con esto antes o no tambien)
% - Se hace una aggregation operation pero la clasificación no suele ser en el espacio hyperbolico

Type inventories for the task of fine-grained entity typing \cite{ling2012fine, yosef2012hyena} have grown in size and complexity \cite{delCorro2015finet, choi2018ultra}.
Researchers have tried to incorporate hierarchical information on the type distribution in different manners \cite{shimaoka2016attentive, ren2016AFET}. 
\newcite{shimaoka2017neural} encode the hierarchy through a sparse matrix. 
\newcite{xuBarbosa2018hierarchyAware} model the relations through a hierarchy-aware loss function.
\newcite{xiong2019inductiveBias} derive a graph from type co-occurrence statistics in the dataset.
Experimental evidence suggests that our model encodes similar hierarchical information without the need to provide it explicitly.

Hyperbolic representations have been employed for Question Answering \cite{tay2018hyperQA}, in Machine Translation \cite{gulcehre2018hyperAttentionNet}, and modeling language \cite{dhingra2018embeddingTextInHS, tifrea2018poincareGlove}. 
We build upon the hyperbolic neural layers introduced in \newcite{ganea2018hyperNN}, and develop the missing components to perform, not binary, but multi-class multi-label text classification.
We test the proposed model not with a synthetic dataset, but on a concrete downstream tasks, such as entity typing. 
Our work resembles \newcite{lopez-etal-2019-fine} and \newcite{chen2019hyperIM}, though they separately learn embeddings for type labels and text representations in hyperbolic space, %in order to facilitate information sharing among them, 
whereas we do it in an integrated fashion.

%\todo{The philosophical (?) framing I mean}{Overall,} we advocate for alternative representation methods, with a well-established mathematical foundation, for efficient and effective neural NLP components, pursuing a similar goal to \newcite{yitay2019lightweightQuat}.

%From "An Attentive Fine-Grained Entity Typing Model with Latent Type Representation": "Fine-grained entity typing is usually formulated as a multi-label classification problem. Previous approaches (Ling and Weld, 2012; Choi et al., 2018; Xin et al., 2018) typically address it with binary relevance that decomposes the problem into isolated binary classification subproblems and independently predicts each type. However, this method is commonly criticized for its label independence assumption, which is not valid for fine- grained entity typing."

%%%%%%%%%%%%%%%%%%%%%%%%%%%%%%%%%%%%%%%%%%%%%%%%%%%%%%%%%%%%
\section{Conclusions}
\label{sec:concl}
% 1) X (+define X if not obvious) is an important problem 
% 2) The core challenges are this and that. 
% 3) Previous work on X has addressed these with Y, but the problems with this are Z. 
% 4) In this work we do W (?). 
% 5) This has the following appealing properties and our experiments show this and that.

Incorporating hierarchical information from the label inventory into neural models has become critical to improve performance.
Hyperbolic spaces are an exciting approach since they are naturally equipped to model hierarchical structures.
However, previous work integrated  isolated components into neural systems. 
In this work we propose a fully hyperbolic model and showcase its effectiveness on challenging datasets. 
Our hyperbolic model automatically infers the latent hierarchy from the class distribution, captures implicit hyponymic relations in the inventory and achieves a performance comparable to state-of-the-art systems on very fine-grained labels with a remarkable reduction of the parameter size. 
This emphasizes the importance of choosing a metric space suitable to the data distribution as an effective inductive bias to capture fundamental properties, such as hierarchical structure.

Moreover, we illustrate ways to integrate different components with Euclidean layers, showing their strengths and drawbacks. An interesting future direction is to employ hyperbolic representations in combination with contextualized word embeddings. We release our implementation with the aim to ease the adoption of hyperbolic components into neural models, yielding lightweight and efficient systems.

%Add future work! En Gulcehre citan un paper y dicen "future work seria hacer lo mismo que CITE, pero co hyperbolic whatever. 
%Para mi future work podría ser explorar variations de HyperbolicMLR para paliar algunas de las desventajas de Softmax (Ver https://arxiv.org/pdf/1812.04616.pdf, recordar RepL4NLP19).
%De Gulcehre: "Similarly as a future work, an interesting potential future direction is to use hyperbolic...", or say clearly that I do not use contextualized word embeddings and future work: explore hyperbolic representation in combination with contextualized word embeddings

\section*{Acknowledgments}
This work has been supported by the German Research Foundation (DFG) as part of the Research Training Group Adaptive Preparation of Information from Heterogeneous Sources (AIPHES) under grant No. GRK 1994/1 and the Klaus Tschira Foundation, Heidelberg, Germany.

% From \cite{becigneul2019riemannianMethods}
% Driven by recent work in learning non-Euclidean embeddings for symbolic data, we propose to

\bibliography{mybib}
\bibliographystyle{acl_natbib}

\newpage
\appendix

\include{appendix}

\end{document}

%% file: appendix.tex
\section{Basics of Riemannian Geometry}
\label{sec:appendix-riemmanian}

\smallskip
\noindent
\textbf{Manifold:} a $n$-dimensional manifold \manifold{} is a space that can locally be approximated by \realto{n}. It generalizes the notion of a 2D surface to higher dimensions. More concretely, for each point $x$ on \manifold{}, we can find a \textit{homeomorphism} (continuous bijection with continuous inverse) between a neighbourhood of $x$ and \realto{n}.

\smallskip
\noindent
\textbf{Tangent space:} the \textit{tangent space} $T_x \mathcal{M}$ at a point $x$ on \manifold{} is a $n$-dimensional hyperplane in \realto{n+1} that best approximates \manifold{} around $x$. It is the first order linear approximation.

\smallskip
\noindent
\textbf{Riemannian metric:} A \textit{Riemannian metric} $g = (g_x)_{x \in \mathcal{M}}$ on \manifold{} is a collection of inner-products $g_x: T_x \mathcal{M} \times T_x \mathcal{M} \rightarrow \mathbb{R}$ varying smoothly with $x$ on tangent spaces. Riemannian metrics can be used to measure distances on manifolds

\smallskip
\noindent
\textbf{Riemannian manifold:} is a pair $($\manifold{}$, g)$, where \manifold{} is a smooth manifold and $g = (g_x)_{x \in \mathcal{M}}$ is a Riemannian metric.

\smallskip
\noindent
\textbf{Geodesics:} $\gamma: [0,1] \rightarrow$ \manifold{} are the generalizations of straight lines to Riemannian manifolds, i.e., constant speed curves that are locally distance minimizing. In the Poincar\'e disk model, geodesics are circles that are orthogonal to the boundary of the disc as well as diameters.

\smallskip
\noindent
\textbf{Parallel transport:} defined as $P_{x \rightarrow y}: T_x \mathcal{M} \rightarrow T_y \mathcal{M}$, is a linear isometry between tangent spaces that corresponds to moving tangent vectors along geodesics. It is a generalization of translation to non-Euclidean geometry, and it defines a canonical way to connect tangent spaces.

%%%%%%%%%%%%%%%%%%%%%%%%%%%%%
\section{M\"obius Operations}
\label{sec:appendix-mob-operations}

\smallskip
\noindent
\textbf{M\"obius scalar multiplication:} for $x \in \mathbb{D}^n \backslash \{0\}$ the M\"obius scalar multiplication by $r \in$ \real{} is defined as:
\begin{equation}
    r \otimes x = \operatorname{tanh}(r \operatorname{tanh}^{-1}(\|x\|)) \frac{x}{\|x\|}
\end{equation}
and $r \otimes 0 := 0$. By making use of the $\operatorname{exp}$ and $\operatorname{log}$ maps, this expression is reduced to:
\begin{equation}
    r \otimes x = \operatorname{exp}_{\textbf{0}}(r \operatorname{log}_{\textbf{0}}(x)), \quad \forall r \in \mathbb{R}, x \in \mathbb{D}^n
\end{equation}

\smallskip
\noindent
\textbf{Exponential and logarithmic maps:} The mapping between the tangent space and hyperbolic space is done by the exponential map $\operatorname{exp}_{x}: T_x \mathbb{D}^n \rightarrow \mathbb{D}^n$ and the logarithmic map $\operatorname{log}_{x}: \mathbb{D}^n \rightarrow T_x \mathbb{D}^n$. They are given for $v \in T_x \mathbb{D}^n \backslash \{0\}$ and $y \in \mathbb{D}^n \backslash \{0\}, y \ne x$: 

\begin{equation}
\small
\begin{aligned}
& \operatorname{exp}_x(v) = x \oplus \left(\operatorname{tanh}\left(\frac{\lambda_x \|v\|}{2} \right) \frac{v}{\|v\|} \right) \\
& \operatorname{log}_x(y) = \frac{2}{\lambda_x} \operatorname{tanh}^{-1}(\|-x \oplus y\|) \frac{-x \oplus y}{\|-x \oplus y\|}
\end{aligned}
\vspace{-2mm}
\end{equation}

These expressions become more appealing when $x = 0$, that is, at the origin of the space. It can be seen that the matrix-vector multiplication formula is derived from $M \otimes y = \operatorname{exp}_{\textbf{0}}(M \operatorname{log}_{\textbf{0}}(y))$. 
The point $y \in \mathbb{D}^n$ is mapped to the tangent space $T_\textbf{0} \mathbb{D}^n$, the linear mapping $M$ is applied in the Euclidean subspace, and finally the result is mapped back into the ball. A similar approach holds for the M\"obius scalar multiplication and the application of pointwise non-linearity functions to elements in the Poincar\'e ball (see \newcite{ganea2018hyperNN}, Section~2.4).

\smallskip
\noindent
\textbf{Parallel transport with $\operatorname{exp}$ and $\operatorname{log}$ maps:} By applying the $\operatorname{exp}$ and $\operatorname{log}$ maps the parallel transport in the Poincar\'e ball for a vector $v \in T_\textbf{0} \mathbb{D}^n$ to another tangent space $T_x \mathbb{D}^n$, is given by:
\begin{equation}
    P_{\textbf{0} \rightarrow x}(v) = \operatorname{log}_{x}(x \oplus \operatorname{exp}_{\textbf{0}}(v)) = \frac{\lambda_{\textbf{0}}}{\lambda_x} v
\end{equation}

This result is used to define and optimize the $a_k = (\lambda_0 / \lambda_{p_k})a_{k}^{\prime}$ in the Hyperbolic MLR (Appendix~\ref{sec:appendix-mlr})

\section{Hyperbolic Gated Recurrent Unit}
\label{sec:appendix-gru}

To encode the context we apply a hyperbolic version of gated recurrent units (GRU) \citep{cho2014gru} proposed in \newcite{ganea2018hyperNN}:
\begin{equation}
\small
\begin{aligned}
& r_t = \sigma \left(log_0(W^r \otimes h_{t-1} \oplus U^r \otimes x_t \oplus b^r) \right) \\
& z_t = \sigma \left(log_0(W^z \otimes h_{t-1} \oplus U^z \otimes x_t \oplus b^z) \right) \\
& \tilde{h_t} = \operatorname{tanh}^{\otimes}((W diag(r_t)) \otimes h_{t-1} \oplus U \otimes x_t \oplus b) \\
& h_t = h_{t-1} \oplus diag(z_t) \otimes (-h_{t-1} \oplus \tilde{h_t})
\end{aligned}
\end{equation}
where $W \in \mathbb{R}^{d_S \times d_S}, U \in \mathbb{R}^{d_S \times n}, x_t \in \mathbb{D}^n$ and $b \in \mathbb{D}^{d_S}$ (superscripts are omitted). $r_t$ is the reset gate, $z_t$ is the update gate, $diag(x)$ denotes a diagonal matrix with each element of the vector $x$ on its diagonal, and $\sigma$ is the $\operatorname{sigmoid}$ function. %, and $\varphi$ is the $\operatorname{tanh}$.

\section{Distance-based Attention}
\label{sec:appendix-attention}

\subsection{Formulation}
In Equation~\ref{eq:midpoint} we calculate the Lorentz factors for each point $x_i$. The Lorentz factors are given by:
\begin{equation}
\gamma(x) = \frac{1}{\sqrt{1 - \|x\|^2}}
\end{equation}

In the case of \citet{gulcehre2018hyperAttentionNet}, the application of the Einstein midpoint (\citealp{ungar2010barycentric}, Theorem 4.4) requires the mapping of the points onto the Klein model. By applying the M\"obius midpoint, we avoid this mapping, and achieve an attention mechanism that operates only in one model of hyperbolic space.

\subsection{Experimental Observations}
To obtain the weights for the attention mechanism, initially Equation~\ref{eq:attn-weights} was given by:
\begin{equation}
\alpha(q_i, k_i) = f(-\beta d_{\mathbb{D}}(q_i, k_i) - c) 
\end{equation}
We experimented with replacing $f$ for $\operatorname{sigmoid}$ and $\operatorname{softmax}$ functions. We found better performance with the latter one. Moreover, empirical observation lead us to remove the $c$ value, since it converged to zero in all experiments. We believe that the biases $b^Q$ and $b^K$ from Equation~\ref{eq:attn-weights} compensate for this $c$.

\begin{figure}[!b]
\small
\centering
\subfloat[Euclidean Space.]{\label{fig:eu-qs}{\includegraphics[width=.47\linewidth,keepaspectratio]{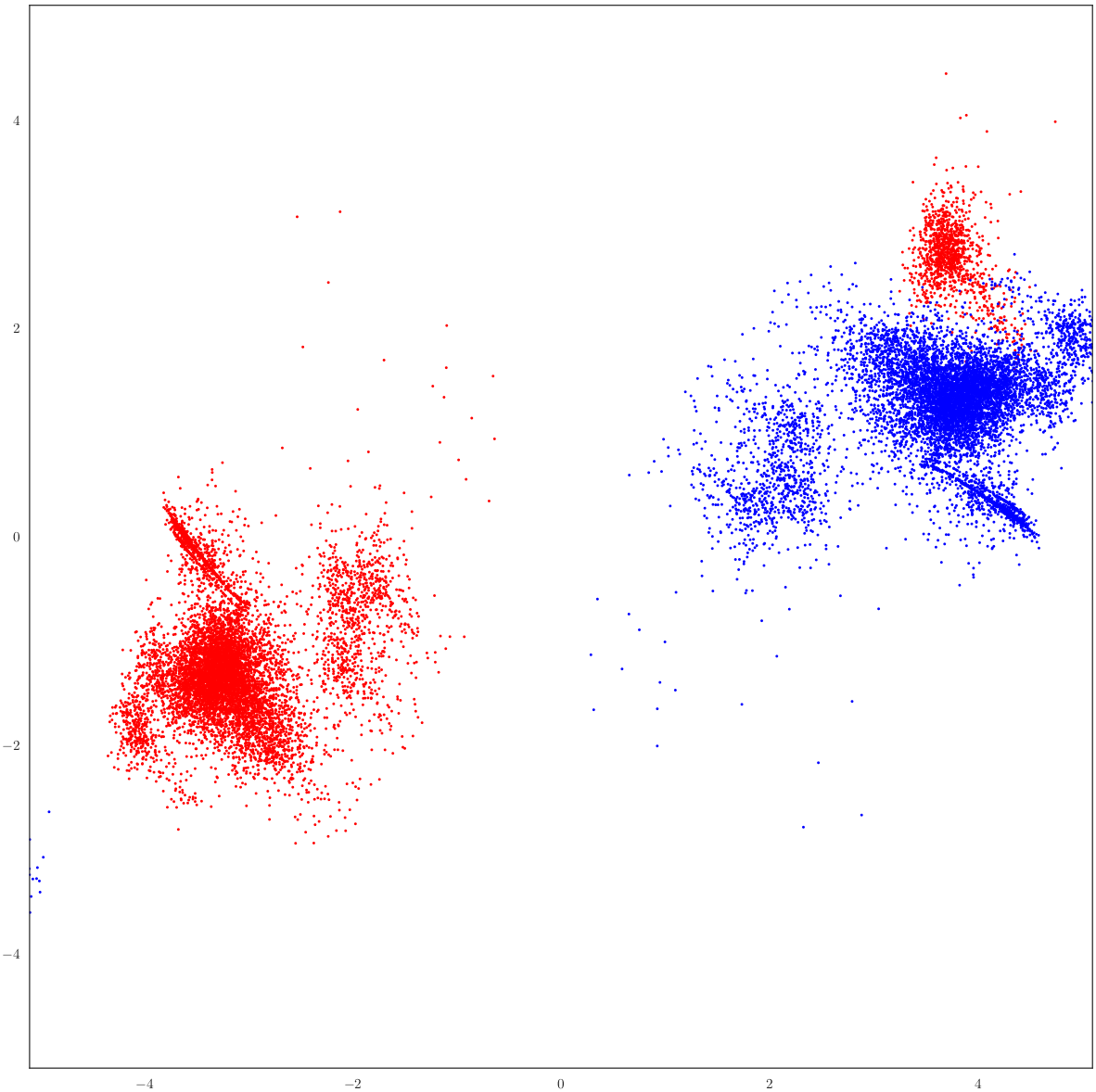}}}\hfill
\subfloat[Hyperbolic Space.]{\label{fig:hy-qs}{\includegraphics[width=.47\linewidth,keepaspectratio]{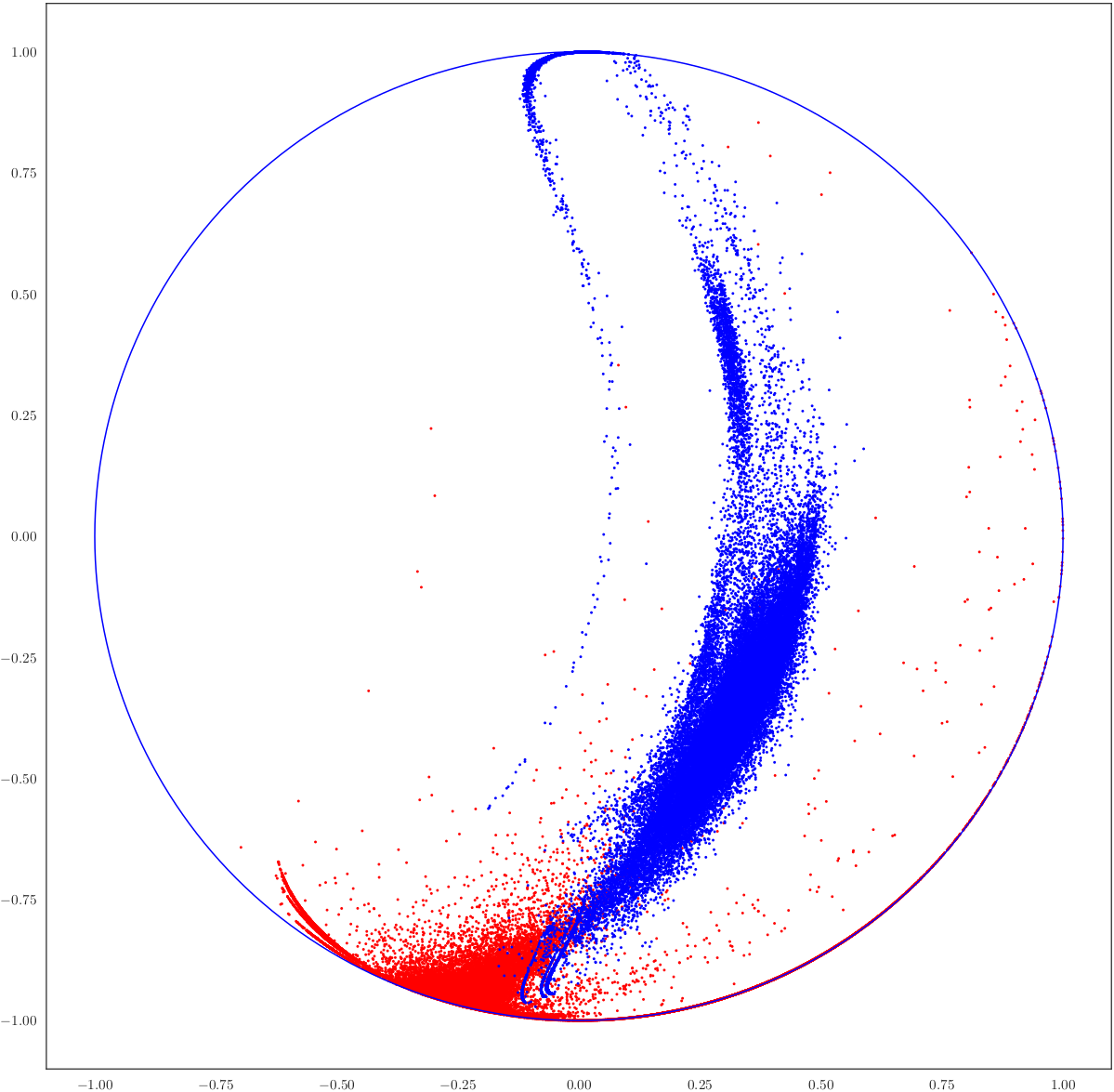}}}
\caption{Queries (red) and keys (blue) projected in 2D for different spaces.}
\label{fig:queries}
\end{figure}

\subsection{Queries and Keys}
To further analyze the attention mechanism we plot the query $q_i$ and key $k_i$ points of Equation~\ref{eq:attn-weights} for both models in Figure~\ref{fig:queries}. It must be recalled that the shorter the distance between points, the higher the attention weight that the word gets assigned. Furthermore, we observed that the attention gets prominently centered on the mention in both models, assigning very low weights on the rest of the words in the context.

In the Euclidean space we can clearly distinguish the two clusters which make the distance-based attention to give very low weights on most words of the context. The small red cluster on the top right of the image belongs to points corresponding to words in the mention span. These words get projected very close to the key vector, in order to minimize the distance and increase the attention weight.

On the hyperbolic model, the queries get clustered at the bottom of the plot, whereas the keys are the points adjusting the distance to define the weight on each word.

\section{Multinomial Logistic Regression}
\label{sec:appendix-mlr}

\subsection{Hyperbolic MLR}
The original formula from \newcite{ganea2018hyperNN} for MLR in the Poincar\'e ball, given $K$ classes and $k \in \{1,...,K\}$, $p_k \in \mathbb{D}^n$, $a_k \in T_{p_k} \mathbb{D}^n \backslash \{0\}$, the formula for the hyperbolic MLR is:
\begin{equation}
\resizebox{.85\hsize}{!}{$
\begin{multlined}
p(y=k|x) \propto \\
f\left(\frac{\lambda_{p_k}^c \|a_k\|}{\sqrt{c}} \operatorname{sinh}^{-1} \left(\frac{2 \sqrt{c} \langle -p_k \oplus x, a_k\rangle}{(1 - c\| -p_k \oplus x \|^2)\|a_k\|} \right) \right)
\end{multlined}
$}
\label{eq:app-mlr}
\end{equation}
Where $x \in \mathbb{D}^n$, $p_k$ and $a_k$ are trainable parameters, and $c$ is a parameter in relation to the radius of the Poincar\'e ball $r = 1 / \sqrt{c}$ which in this work we assume to be $c = 1$, hence it is omitted of the formulations. 
Since $a_k \in T_{p_k} \mathbb{D}^n$ and therefore depends on $p_k$, it is unclear how to perform optimization. The solution proposed by \newcite{ganea2018hyperNN} is to re-express it as: 
\begin{equation}
a_k = P_{\textbf{0} \rightarrow p_k}(a_{k}^{\prime}) = \frac{\lambda_{\textbf{0}}}{\lambda_{p_k}} a_{k}^{\prime}    
\end{equation}
where $a_{k}^{\prime} \in T_{\textbf{0}} \mathbb{D}^n = \mathbb{R}^n$. In this way we can optimize $a_{k}^{\prime}$ as a Euclidean parameter.
Finally, when we use $a_k^{\prime}$ instead of $a_k$, the formula for the MLR is:
\begin{equation}
\resizebox{.85\hsize}{!}{$
\begin{multlined}
p(y=k|x) \propto \\
f\left(2 \|a_k^{\prime}\| \operatorname{sinh}^{-1} \left(\frac{2 \langle -p_k \oplus x, a_k^{\prime}\rangle}{(1 - \| -p_k \oplus x \|^2)\|a_k^{\prime}\|} \right) \right)
\end{multlined}
$}
\end{equation}

\subsection{Euclidean MLR}
The Euclidean formulation of the MLR is given by: 
\begin{equation}
    p(y=k|x) \propto f(4\langle -p_k \oplus x, a_k\rangle)
\end{equation}
This equation arise from taking the limit of $c \rightarrow 0$ in Equation~\ref{eq:app-mlr}. In that case, $f(4\langle -p_k \oplus x, a_k\rangle) = f((\lambda_{p_k}^0)^2\langle -p_k \oplus x, a_k\rangle) = f(\langle -p_k \oplus x, a_k\rangle_0)$.

\section{Experimental Details}
\label{sec:appendix-hyperparams}

%\smallskip
%\noindent
%\textbf{Setup:} Given the dimension of the model $d$, we set the dimension of the mention projection to $d_M = 2d$.
For the context-GRU we use $\operatorname{tanh}$ as non-linearity to establish a fair comparison against the classical GRU \citep{cho2014gru}. On the char-RNN we use the $\operatorname{identity}$ (no non-linearity).
The MLR is fed with the final representation achieved by the concatenation of mention and context features: $\operatorname{concat}(\mathbf{M}, \mathbf{C}, \mathbf{S}) \in \mathbb{D}^{m}$ with $m = d_M + d_C + 2 d_S$.

In the \textsc{xLarge} model, we use the Euclidean encoder in all experiments given time constraints.

\smallskip
\noindent
\textbf{Hyperparameters:} Both hyperbolic and Euclidean models were trained with the hyperparameters detailed in Table~\ref{tab:hyperparams}.

\smallskip
\noindent
\textbf{Dropout:} We apply low values of dropout given that the model was very sensitive to the this parameter. We consider this a logical behaviour since the distances in hyperbolic space grow exponentially with the norm of the points, making the model very responsive to this parameter.

\smallskip
\noindent
\textbf{Numerical Errors:} they appear when the norm of the hyperbolic vectors is very close to $1$ or $0$. To avoid them we follow the recommendations reported on \newcite{ganea2018hyperNN}. The result of hyperbolic operations is always projected in the ball of radius $1 − \epsilon$, where $\epsilon = 10^{−5}$. When vectors are very close to $0$, they are perturbed with an $\varepsilon = 10^{−15}$ before they are used in any of the above operations. Finally, arguments of the $tanh$ function are clipped between $\pm15$, while arguments of $tanh^{−1}$
are clipped in the interval $[-1 + 10^{−15}, 1 − 10^{−15}]$. Finally, and by recommendations of the Geoopt developers \cite{geoopt2019kochurov}, we operate on floating point of $64$ bits.

\smallskip
\noindent
\textbf{Initialization:} we initialize character and positional embeddings randomly from the uniform distribution $U(-0.0001, 0.0001)$. In the case of the hyperbolic model, we map them into the ball with the $\operatorname{exp}_{\textbf{0}}$ map.
We initialize all layers in the model using \textit{Glorot uniform initialization}.

\begin{table}[!h]
\centering
\begin{tabular}{lr}
\toprule
\textbf{Parameter} & \multicolumn{1}{l}{\textbf{Value}} \\
\midrule
Batch size \textsc{base} & $900$ \\
Batch size \textsc{large} & $350$ \\
Batch size \textsc{xLarge} & $160$ \\
\textsc{base} $d_M$ & 40 \\
\textsc{base} $d_C$ & 20 \\
\textsc{base} $d_S$ & 20 \\
\textsc{base} $d_M + d_C + 2d_S$ & 100 \\
\textsc{large} $d_M$ & 100 \\
\textsc{large} $d_C$ & 50 \\
\textsc{large} $d_S$ & 50 \\
\textsc{large} $d_M + d_C + 2d_S$ & 250 \\
\textsc{xLarge} $d_M$ & 200 \\
\textsc{xLarge} $d_C$ & 100 \\
\textsc{xLarge} $d_S$ & 100 \\
\textsc{xLarge} $d_M + d_C + 2d_S$ & 500 \\
Mention non-linearity & $\operatorname{tanh}$ \\
Context non-linearity & $\operatorname{tanh}$ \\
Epochs & $40$ \\
Crowd cycles & $5$ \\
Input dropout & $0.2$ \\
Concat dropout & $0.1$ \\
Learning rate & $0.0005$ \\
Weight decay & $0.0$ \\
Max. gradient norm & $5$ \\
\bottomrule
\end{tabular}
\caption{Hyperparameters of the models.}
\label{tab:hyperparams}
\end{table}

\smallskip
\noindent
\textbf{Exponential and logarithmic map:} In the case of the Glove embedding ablation (Section~\ref{sec:embed-ablation}), we used the $100d$ version, trained over Wikipedia and Gigaword\footnote{\url{http://nlp.stanford.edu/data/glove.6B.zip}}. By directly applying the logarithmic map, the embeddings were projected close to the border of the ball, making the model very unstable. To overcome this, we use a parameter $c$ described in \newcite{ganea2018hyperNN} to adjust the radius of the ball, which helps to project the embeddings closer to the origin of the space.

\smallskip
\noindent
\textbf{Hardware:} All experiments for the hyperbolic and Euclidean models were performed using 2 NVIDIA P40 GPUs, with the batch sizes specified in Table~\ref{tab:hyperparams}.

\section{Closest Types}
\label{sec:appendix-closest-pk}
We report the points $p_k$ learned by the model to define the hyperplanes of Equation~\ref{eq:mlr}. Table~\ref{tab:appendix-closest-pk} shows the types corresponding to the closest points, measured by their hyperbolic distance $d_{\mathbb{D}}$ (see Eq~\ref{eq:hyper-dist}), to the \textit{coarse} labels. We observe that the types are highly correlated given that they often co-occur in the same context.

\begin{figure}[!t]
\centering
\includegraphics[width=0.9\linewidth,keepaspectratio]{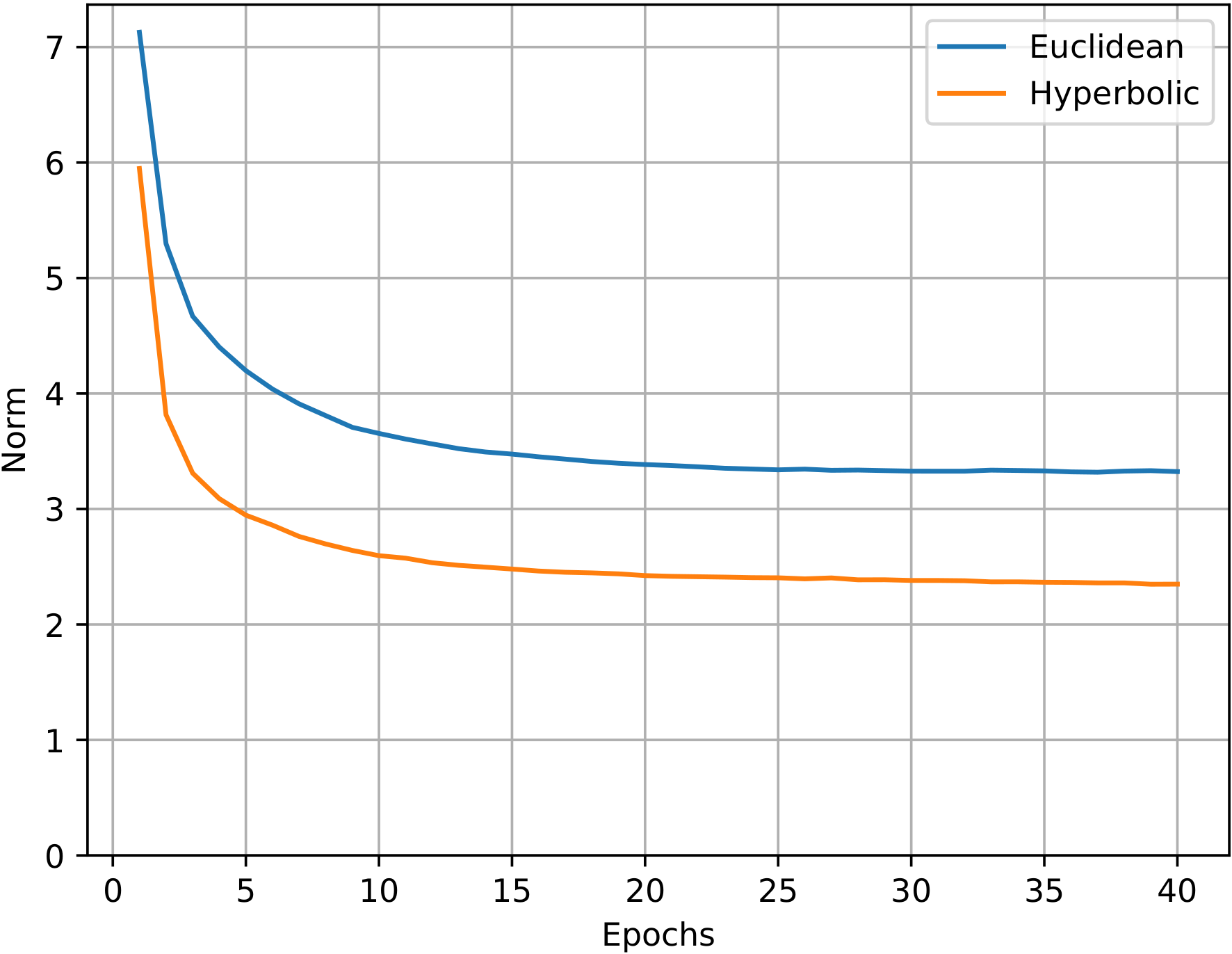}
\caption{Norm of \textit{text vectors} for the Euclidean and hyperbolic model. The hyperbolic norm is measured as the hyperbolic distance $d_{\mathbb{D}}$ from the origin to the point, hence the values can be greater than $1$.}
\label{fig:text-norms}
\end{figure}

\section{More Experimental Observations}
\label{sec:observations}

\paragraph{Text vectors norms:} By ``text vector'' we refer the concatenated vector of the context, mention and char-level mention representations before the MLR layer. We report the average norm of this vectors per training epoch, for the \textsc{20D} Euclidean and hyperbolic model on Figure~\ref{fig:text-norms}. The norm of the vectors of the hyperbolic model are measured according to the hyperbolic distance $d_{\mathbb{D}}$ (see Equation~\ref{eq:hyper-dist}). That is, we take the hyperbolic distance from the origin to the point, thus the values are above one. The norm of the Euclidean model is measured according to the Euclidean norm. We observe that both models learn to reduce the norm of the vectors, and it is noticeable that the convergence value for the Euclidean model is higher than for the hyperbolic model.

\begin{table*}[!t]
\centering
\adjustbox{max width=\textwidth}{
\begin{tabular}{llllll|llllll}
\toprule
\multicolumn{2}{c}{\texttt{organization}} & \multicolumn{2}{c}{\texttt{institution}} & \multicolumn{2}{c}{\texttt{firm}} & \multicolumn{2}{c}{\texttt{group}} & \multicolumn{2}{c}{\texttt{unit}} & \multicolumn{2}{c}{\texttt{division}} \\
\textbf{Types} & $d_{\mathbb{D}}$ & \textbf{Types} & $d_{\mathbb{D}}$ & \textbf{Types} & $d_{\mathbb{D}}$ & \textbf{Types} & $d_{\mathbb{D}}$ & \textbf{Types} & $d_{\mathbb{D}}$ & \textbf{Types} & $d_{\mathbb{D}}$ \\
\midrule
institution & 0.34 & firm & 0.24 & business & 0.23 & unit & 0.34 & division & 0.26 & subsidiary & 0.25 \\
company & 0.35 & company & 0.26 & institution & 0.24 & gathering & 0.34 & theatre & 0.28 & unit & 0.26 \\
news\_agency & 0.36 & university & 0.26 & company & 0.25 & subject & 0.34 & activist & 0.28 & track & 0.28 \\
business & 0.38 & operator & 0.28 & maker & 0.27 & administration & 0.36 & position & 0.28 & half & 0.28 \\
administration & 0.40 & maker & 0.28 & operator & 0.28 & affiliation & 0.36 & half & 0.28 & activist & 0.29 \\
\bottomrule
\multicolumn{2}{c}{\texttt{location}} & \multicolumn{2}{c}{\texttt{state}} & \multicolumn{2}{c}{\texttt{country}} & \multicolumn{2}{c}{\texttt{place}} & \multicolumn{2}{c}{\texttt{space}} & \multicolumn{2}{c}{\texttt{half}} \\
\textbf{Types} & $d_{\mathbb{D}}$ & \textbf{Types} & $d_{\mathbb{D}}$ & \textbf{Types} & $d_{\mathbb{D}}$ & \textbf{Types} & $d_{\mathbb{D}}$ & \textbf{Types} & $d_{\mathbb{D}}$ & \textbf{Types} & $d_{\mathbb{D}}$ \\
\midrule
state & 0.33 & country & 0.29 & state & 0.31 & space & 0.40 & half & 0.28 & peak & 0.26 \\
cemetery & 0.35 & half & 0.31 & nation & 0.31 & localization & 0.40 & shopping\_mall & 0.29 & operator & 0.26 \\
space & 0.35 & agency & 0.31 & agency & 0.32 & place\_name & 0.40 & venue & 0.29 & theatre & 0.26 \\
half & 0.35 & activist & 0.32 & kingdom & 0.34 & close & 0.41 & landmark & 0.30 & placement & 0.26 \\
area & 0.36 & unit & 0.32 & world & 0.35 & birthplace & 0.41 & localization & 0.30 & summit & 0.26 \\
\bottomrule
\multicolumn{2}{c}{\texttt{event}} & \multicolumn{2}{c}{\texttt{conflict}} & \multicolumn{2}{c}{\texttt{war}} & \multicolumn{1}{c}{\texttt{time}} & \multicolumn{1}{c}{} & \multicolumn{2}{c}{\texttt{duration}} & \multicolumn{2}{c}{\texttt{calendar}} \\
\textbf{Types} & $d_{\mathbb{D}}$ & \textbf{Types} & $d_{\mathbb{D}}$ & \textbf{Types} & $d_{\mathbb{D}}$ & \textbf{Types} & $d_{\mathbb{D}}$ & \textbf{Types} & $d_{\mathbb{D}}$ & \textbf{Types} & $d_{\mathbb{D}}$ \\
\midrule
conflict & 0.44 & war & 0.34 & guerrilla & 0.32 & duration & 0.40 & calendar & 0.30 & date & 0.22 \\
activist & 0.45 & dispute & 0.36 & conflict & 0.34 & period & 0.43 & peak & 0.31 & phrase & 0.25 \\
election & 0.45 & series & 0.37 & military & 0.35 & length & 0.46 & half & 0.32 & second & 0.26 \\
activity & 0.46 & guerrilla & 0.38 & citizen & 0.36 & month & 0.46 & second & 0.32 & activist & 0.27 \\
holiday & 0.46 & future & 0.38 & situation & 0.36 & date & 0.46 & fantasy & 0.32 & need & 0.28 \\
\bottomrule
\multicolumn{2}{c}{\texttt{object}} & \multicolumn{2}{c}{\texttt{machine}} & \multicolumn{2}{c}{\texttt{computer}} & \multicolumn{2}{c}{\texttt{entity}} & \multicolumn{2}{c}{\texttt{separation}} & \multicolumn{2}{c}{\texttt{placement}} \\
\textbf{Types} & $d_{\mathbb{D}}$ & \textbf{Types} & $d_{\mathbb{D}}$ & \textbf{Types} & $d_{\mathbb{D}}$ & \textbf{Types} & $d_{\mathbb{D}}$ & \textbf{Types} & $d_{\mathbb{D}}$ & \textbf{Types} & $d_{\mathbb{D}}$ \\
\midrule
machine & 0.37 & computer & 0.29 & version & 0.29 & separation & 0.43 & placement & 0.27 & position & 0.25 \\
arrangement & 0.39 & theatre & 0.30 & machine & 0.30 & relative & 0.44 & missionary & 0.27 & localization & 0.26 \\
medium & 0.39 & operator & 0.30 & communication & 0.30 & meaning & 0.44 & meaning & 0.27 & half & 0.26 \\
method & 0.39 & card\_game & 0.31 & activist & 0.31 & warlord & 0.45 & variation & 0.27 & separation & 0.27 \\
representation & 0.39 & core & 0.31 & maker & 0.32 & baseball & 0.45 & phrase & 0.27 & winner & 0.27 \\
\bottomrule
\end{tabular}
}
\caption{Closest $p_k$ points in the Poincar\'e Ball to \textit{coarse} entity types, with their hyperbolic distance. In many cases, a hierarchical relation holds with the closest type. For example: \texttt{firm} \textit{is-a} \texttt{institution} \textit{is-a} \texttt{organization}.}
\label{tab:appendix-closest-pk}
\end{table*}